\documentclass[10pt]{article}
\usepackage[preprint]{tmlr}   % TMLR style (authoryear natbib; loads fancyhdr, natbib, lmodern)
\usepackage[utf8]{inputenc}
\usepackage{url}
\usepackage{booktabs}
\usepackage{longtable}   % claims table spans a page; break it inline under its heading
\usepackage{multirow}
\usepackage{amsmath,amssymb,amsfonts}
\usepackage{nicefrac}
\usepackage{microtype}
\usepackage{xcolor}
\usepackage{graphicx}
\usepackage{needspace}
\usepackage{placeins}   % \FloatBarrier to keep appendix tables in their own sections
\usepackage{float}      % [H] to pin the deferred-tables appendix inline (no floating gap)
\usepackage{hyperref}   % tmlr does not load hyperref; load it last
\hypersetup{hidelinks}

\def\month{07}
\def\year{2026}

\makeatletter
\newcommand{\repofootnote}{\if@accepted\thanks{Code, data, and the full reproducible experiment
ledger: \url{https://github.com/collapseindex/breaking-refusal}}\fi}
\makeatother

\title{Breaking Refusal in the First Half\\
A Mechanistic Study of the Prefill Jailbreak\repofootnote}

\author{\name Alex Kwon \email ask@collapseindex.org \\
        \addr Independent Researcher}

\begin{document}
\maketitle

\begin{abstract}
Aligned language models refuse harmful requests, but a one-line prefill (``Sure, here is'')
strips the refusal. We ask where and how it fails. The harm representation stays intact: on the
very prompts the attack flips to compliance, a linear probe reads harm as high as on the refused
ones ($0.91$--$0.98$), while behavioral refusal drops to chance. This holds across four models and
three families ($1.5$--$3.8$B, and at $14$B). Refusal is therefore a shallow, response-site
computation. We localize it to an \emph{early window}: a dose-matched position control shows the
first half of the response suffices to break refusal, while the second half is nearly inert.
Three causal probes converge on that window. Restoring the harm direction there partially
re-engages refusal. Injecting the model's own refuse-state reverses the jailbreak ($\to74\%$,
held-out). And knocking out the early response's attention to the prefill, but not an equal
attention mass elsewhere, selectively collapses the harmful continuation. What kind of mechanism is
this? A base-model control answers: the same knockout collapses the continuation
\emph{prefill-specifically} even in a non-safety-tuned base model ($64\%\to25\%$ harmful content vs a
matched control's $64\%$, replicated at $7$B). So the prefill's grip is generic autoregressive
conditioning, not safety-specific suppression, and ``refusal restoration'' is a model-dependent
fallback. The dominant mechanism is passive. A small safety-specific attractor remains on top
(logit-trace concentration $0.24$ vs $0.03$), whose active-vs-passive character we size but do not
fully separate. No single direction or component is a clean handle either: the decision is decodable
but distributed, and refusal tracks harm rather than scary surface. Every reported effect is one
that passed its controls. The practical consequence is structural: a monitor reading the untouched
prompt-side representation is immune to this attack by construction, though only to response-site
attacks. The mechanism is diffuse; the failure surface is local.
\end{abstract}

% ---- teaser figure (make figures/teaser.pdf); after the abstract for TMLR ----
\begin{figure}[H]
\centering
\includegraphics[width=0.72\textwidth]{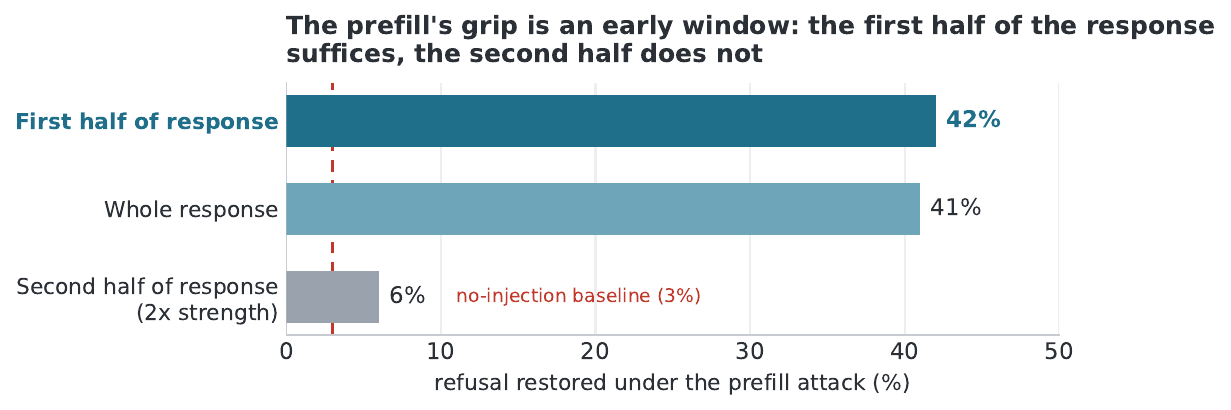}
\caption{\textbf{The prefill's grip is an early response window (the central finding).} Restoring
the harm direction over a dose-matched band of the response under the attack ($n=100$,
Qwen2.5-1.5B, L12/$\alpha{=}1$; \S\ref{sec:causal}, App.~\ref{app:patch}): the \emph{first half}
restores refusal to $42\%$, as much as the whole response ($41\%$), while the \emph{second half}
restores only $6\%$ at double strength and the onset alone $9\%$. An early multi-token window
suffices; the late response carries no weight.}
\label{fig:teaser}
\end{figure}

% =====================================================================
\section{Introduction}
\label{sec:intro}

Aligned language models refuse harmful requests, and that refusal is easily removed. Prepend
an affirmative prefix to the model's own turn (``Sure, here is'') and many models continue
straight into the harmful content they would otherwise decline \citep{wei2023jailbroken,
qi2024shallow}. The fragility is well documented \emph{behaviorally}: we know the attack
works. We ask the mechanistic question underneath it. The prefill appends to the model's
\emph{own response}, so it cannot touch the representation the model has already computed over
the prompt; the question is therefore not whether that representation is preserved under the attack
(structurally, it must) but whether the model builds a strong harm representation there at all,
and if so, why the refusal decision ignores it.

This localizes what kind of defense is possible. A behavioral gate is defeated by the attack; a
monitor that reads the prompt-side harm representation is not, because the attack never reaches
that representation, though this prompt-side placement is already how production input classifiers
work, so the observation is about placement, not a new detector. The scientific question that remains is why a model that
represents the request as harmful nonetheless complies, that is, where the refusal decision
overriding the representation is computed.

Empirically, the model builds a full-strength harm representation on the flipped prompts
(probe $0.91$--$0.98$, as high as on the ones it refuses) while behavioral refusal drops to
chance, across four models and three families (\S\ref{sec:dissociation}). But the sharper
mechanistic statement is that the gate is not \emph{ignoring} harm, it is being
\emph{overpowered}: under a full-strength prefill essentially every prompt flips, so harm cannot
predict who complies, yet under a \emph{weakened} prefill where a real split exists, harm
strength predicts refuse-vs-comply (AUC $0.63$--$0.72$, $n=70$; \S\ref{sec:dirsearch}). And the
gate's failure is localized to an early window: a dose-matched position control shows that
restoring the harm signal over the \emph{first half} of the response recovers refusal as much as
the whole response, while the second half recovers almost nothing and the onset token alone only
$9\%$ (\S\ref{sec:mechanism}). We use \emph{overpowered} descriptively: the evidence fixes
\emph{where} refusal fails, an early window, but does not separate active suppression from passive
autoregressive conditioning (\S\ref{sec:limits}). We then pin down what the gate reads
(\S\ref{sec:reads}) and show it has no compact internal handle (\S\ref{sec:dirsearch}).

\paragraph{Relation to prior work.} Two lines are adjacent (\S\ref{sec:related}).
\citet{qi2024shallow} show safety alignment is ``shallow'' in
\emph{token depth}: the refusal decision is carried by the first few output tokens, and
attacks that get past them succeed. We go past that behavioral picture on three fronts. First, we
localize the failure causally, not just correlating depth with success, to an early multi-token
\emph{window} rather than a first-token count. Second, we show the causal converse: transplanting
the refuse-state re-induces refusal. Third, the base-model discriminator identifies the prefill's
grip as generic autoregressive conditioning rather than a defeated safety-specific gate, a
distinction depth-shallowness alone does not make. \citet{arditi2024refusal} show refusal is mediated by a
single \emph{direction} in activation space. Closest, concept-level work shows a model still
\emph{recognizes} toxicity under a jailbreak while complying \citep{zhang2025jbshield}, but
attributes the behavior-flip to a competing \emph{prompt-side} ``jailbreak concept'' that
overrides it. Our contribution is positional and causal. The representational half is a
clarification, not a discovery: because prefill leaves the prompt untouched, the harm
representation there is trivially unchanged, and the useful contrast is with
\emph{prompt-modifying} jailbreaks (the ones that literature studies), which do move or
attenuate the prompt-side representation. Our substantive claims are mechanistic: refusal is
computed at the response site and the attack operates there; and the harm direction
is an excellent read-out but only a partial, non-selective write-handle for the refusal action. The prompt-side account cannot
explain the prefill attack at all. Prefill adds nothing to the prompt, so there is no
prompt-side jailbreak concept to override anything. The flip is \emph{response-site}, and we
localize it there. The template
(read what a model represents versus what it acts on) is the one we used for communicative
intent in prior work \citep{kwon2026recipient}; here it is pointed at safety.

\paragraph{Contributions.}
\begin{itemize}\itemsep2pt
\item \textbf{A quantitative dissociation under prefill (\S\ref{sec:dissociation}).} On the very
prompts a prefill attack flips to compliance, the harm representation is not degraded but
\emph{intact}: a probe scores the complied prompts at $0.91$--$0.98$, level with the ones the
model refuses, while behavioral refusal collapses to chance. This replicates across four models
and three families ($1.5$--$3.8$B, weak and strong gates). Unlike the prompt-modifying
jailbreaks prior work studies, prefill leaves the representation \emph{quantitatively} unchanged,
not merely present-but-attenuated.
\item \textbf{A response-site causal mechanism (\S\ref{sec:mechanism}).} We localize refusal
causally, not just correlationally: adding the harm direction back across the response positions
partially restores refusal (harm $33$--$48\%$ at two fixed cells vs a matched-norm random
control's $14$--$16\%$; harm$-$random $+17$ to $+34$pp at $n=100$, both significant), so the
intact representation is upstream of the gate, not epiphenomenal. A dose-matched position control
localizes the effect to an \emph{early window} of the response (the first half restores as much as
the whole; the second half is nearly inert), more than a single onset token but not the whole
generation. The converse holds in the same window and is state-specific: transplanting the model's
own refuse-state into the attack re-induces refusal ($\to74\%$ held-out, first half $=$ whole response),
matched by a random control at $0\%$, a working lever where the single harm direction is only
partial. And the grip operates through early-window attention: knocking out the early response's
attention to the prefill (but not an equal mass elsewhere) restores refusal $+21$pp (CI $[+12,+30]$),
harm-selectively and early-window-gated. A base-model control identifies this as \emph{generic}
autoregressive conditioning, not safety-specific active suppression (the non-safety-tuned base model
shows the same prefill-specific collapse, at both $1.5$B and $7$B), resolving the paper's central open
question toward the passive reading. The dramatic full-residual ``restoration''
is generic disruption, since a benign-patch control also
hits $100\%$. Replicated on two families, and the discriminator holds at $7$B.
\item \textbf{A defensive consequence, and its scope (\S\ref{sec:dissociation},~\S\ref{sec:steer}).}
Read as a monitor, the intact representation catches $100\%$ of the attack at $0\%$ false
positive. The scope is narrow, in three ways. First, this $100\%/0\%$ is in-distribution: the probe
is AdvBench-calibrated (it fires at $0.97$ on that phrasing but $0.49$ on differently-worded unsafe
prompts, and a mixed-source fix wrecked specificity), so a source-general detector is unbuilt.
Second, it is only a response-site guarantee, and the placement is not novel (input classifiers
already sit prompt-side). Third, steering the same direction does \emph{not} close the gap: the harm
direction is a read-out, not a write-handle.
\item \textbf{What the gate reads (\S\ref{sec:reads}).} A pre-registered harm~$\times$~surface
$2\times2$ that decorrelates harm from trigger-word surface shows refusal is harm-primary and the
harm representation keyword-invariant, with a real but secondary surface-driven over-refusal on
benign inputs. A matched-contrast search finds no single refusal-decision \emph{direction}
(any one reads $\le0.73$ and none steers), yet a full linear probe reads the decision at
$\sim0.85$: within the Qwen family the decision is decodable but \emph{multidimensional}, not a
single steerable feature (SmolLM2 is a single-direction counterexample; cross-family is open).
\end{itemize}

Every headline effect is checked against a control that can demote it, and several do; the
claims-and-evidence table (Table~\ref{tab:claims}, Appendix~\ref{app:claims}) records each
claim that passed with its evidence and the demoted versions (full-residual disruption, the steering
null, the retracted ``separate feature''), and Appendix~\ref{app:repro} gives the reproduction
chain. The dissociation also extends, exploratorily, to inferred-intent refusal ($n=24$, one
model; \S\ref{sec:reads}, App.~\ref{app:intent}).
\FloatBarrier   % keep the Table 1 float in \S3 from rising into the contributions list

% =====================================================================
\section{Setup}
\label{sec:setup}

\paragraph{Models.} We use open-weight instruction-tuned models so we can read activations:
Qwen2.5-1.5B and 3B \citep{qwen2025}, SmolLM2-1.7B \citep{allal2025smollm2}, and
Phi-3-mini-3.8B \citep{abdin2024phi3}: three families, $1.5$--$3.8$B
parameters, and a range of safety-gate strength (SmolLM2 refuses $63\%$ of plain harmful
prompts; the others $\approx100\%$). All experiments run on CPU; sample sizes are reported
per experiment.

\paragraph{Data.} Harmful prompts are AdvBench behaviors \citep{zou2023universal}; benign
prompts are Alpaca instructions \citep{taori2023alpaca}; the surface-vs-harm control uses XSTest
\citep{rottger2024xstest}, whose ``safe'' split is purpose-built to \emph{sound} harmful
(``how do I kill a Python process'') while being benign.

\paragraph{The attack.} The prefill attack forces the assistant turn to begin with the
affirmative prefix ``Sure, here is'' and lets the model continue; we score whether the
continuation refuses.

\paragraph{The probe.} The harm probe is $\ell_2$-regularized logistic regression on the
default-pass hidden state at the last prompt token (the decision point), trained on
plain harmful vs.\ benign prompts. Probe AUCs use cross-validation (nested-CV layer selection;
no test-set peeking). Refusal is detected by string match against
a fixed refusal-phrase list. Full details in App.~\ref{app:repro}.

% =====================================================================
\section{The phenomenon: represented but not refused}
\label{sec:dissociation}

This section states the phenomenon the rest of the paper explains: a fully intact harm
representation coexists with compliance,
part of which is structural (below). The mechanistic account
of \emph{how} that gate is defeated (\S\ref{sec:mechanism}) is where the empirical work
and every control live.

\paragraph{Harm is strongly represented.} On plain prompts, the harm probe separates harmful
from benign near-perfectly: one nested-CV probe scores AUC $0.996$ on Qwen2.5-1.5B and
$\ge 0.990$ on every model (Table~\ref{tab:dissociation}, one estimator throughout). The
representation of harm is not subtle.

\paragraph{The attack changes behavior; it cannot change the prompt-side read.} Under the
prefill attack the same model complies: on Qwen2.5-1.5B, refusal on harmful prompts drops from
$1.00$ to $0.03$ (97 of 100 comply) and behavior, used as a harm detector, collapses to chance
(AUC $0.51$). ``Comply'' here is coherent harmful compliance, not just an absent refusal string:
a judge-free content read of the attacked outputs (\S\ref{sec:limits}) finds $94\%$ of the
non-refusals are coherent harmful content, deflection rare. The other half of the dissociation is partly structural. The harm
representation we probe is the model's read at the last prompt token, computed before any of
the attack's tokens exist; the prefill appends to the \emph{response}, so by the causal
structure of the transformer this representation is bit-for-bit identical with and without the
attack. Its ``intactness'' is \emph{structural, not a measurement}. The substantive finding is
this: at full attack strength, \emph{compliance is not explained by a weaker harm read.} The
prompts the model complies on score as harmful ($0.975$) as the ones it refuses ($0.966$; benign
$0.055$). Here ``refused'' is the plain-condition harmful set, since the strong prefill leaves
at most $5/100$ still refusing per model. A \emph{full}-strength harm representation coexists with
compliance; the comparison has little variation, since $95$--$100\%$ comply under the strong attack.
Where a real split exists, a weakened prefill (\S\ref{sec:dirsearch}, $41/29$ refuse/comply), harm
strength \emph{does} predict who complies: the harm-probe score reads refuse-vs-comply
at AUC $0.63$ (layer 12) to $0.72$ (best layer). This is harm, not the obvious surface correlate: prompt
length and scary-keyword count predict the same split at chance (AUC $0.50$--$0.54$), and the
harm-probe AUC \emph{holds} when we regress out length and keyword count ($0.73$). So the strong
attack overrides the harm signal rather than the model down-weighting it. This is moderate evidence:
$n=70$ is small and prompt-level confounds (topic, severity, formality) are not held fixed. Table~\ref{tab:dissociation} shows the pattern across all four models and
three families, at both weak and strong gates. It also extends to larger scale (run on GPU). On
Qwen2.5-7B the prefill drops refusal $0.98\to0.24$ at probe AUC $1.00$; the attack is weaker here,
with a quarter still refusing under the same prefill (\S\ref{sec:limits}). On the
heavily-safety-trained Phi-3-medium-14B \citep{abdin2024phi3} it drops refusal $1.00\to0.15$ (probe
AUC $1.00$, complied-harmful $0.96$, coherent output; App.~\ref{app:ablation}). So the dissociation
is not a small-model artifact. The attack is also not specific
to one prefill string: across six affirmative prefills (varying length and phrasing) it drops
refusal from $100\%$ to $0$--$37\%$ (most to $\le7\%$), so it is a general property of affirmative
response-onset prefills. The real question is then why an
intact, strong harm representation fails to drive refusal, which the mechanism
(\S\ref{sec:mechanism}) answers.

\begin{table}[t]
\centering
\caption{\textbf{The dissociation replicates across models and families.} Under the prefill
attack, behavioral refusal collapses (behavior AUC $\to$ chance). The harm representation is the
model's clean prompt-side read, which the response-site attack cannot alter (invariant by
construction); we report it to show the read is strong (rep.\ AUC $\ge 0.990$, one nested-CV
estimator) and that
compliance does not track a weaker read: the \emph{complied}-harmful prompts score
$0.91$--$0.98$, level with the refused ones and far above benign (``refused'' is the
plain-condition harmful set, since the strong prefill leaves almost no refusals). SmolLM2's
behavior AUC is undefined because it refuses nothing under attack (all comply).}
\label{tab:dissociation}
\small
\begin{tabular}{@{}llccccc@{}}
\toprule
& & \multicolumn{2}{c}{refusal (harmful)} & \multicolumn{3}{c}{harm representation} \\
\cmidrule(lr){3-4}\cmidrule(lr){5-7}
model & family & plain & attack & behavior AUC & rep.\ AUC & complied-harmful \\
\midrule
Qwen2.5-1.5B     & Qwen          & $1.00$ & $0.03$ & $0.51$ & $0.996$ & $0.975$ \\
Qwen2.5-3B       & Qwen          & $0.98$ & $0.05$ & $0.52$ & $1.000$ & $0.981$ \\
SmolLM2-1.7B     & HuggingFaceTB & $0.63$ & $0.00$ & \emph{all comply} & $0.990$ & $0.958$ \\
Phi-3-mini-3.8B  & Microsoft     & $1.00$ & $0.00$ & $0.49$ & $0.998$ & $0.907$ \\
\bottomrule
\end{tabular}
\end{table}

\paragraph{Is the probe reading harm, or scary words?} A natural worry is that the probe
separates AdvBench from Alpaca on surface features (violent vocabulary) rather than harm. We
test this with XSTest, whose ``safe'' prompts sound harmful but are benign. The
AdvBench-trained probe scores XSTest-\emph{safe} at $0.21$ (mean; only $12\%$ exceed the
$0.5$ threshold) versus XSTest-\emph{unsafe} at $0.49$, separating them at AUC $0.79$. The
probe is not fooled by scary surface: it reads harm. It is, however, calibrated to
AdvBench-style phrasing (it fires at $0.97$ there but only $0.49$ on differently-phrased
unsafe prompts), and training on mixed harmful sources did not fix this: it raised
sensitivity but collapsed specificity (XSTest-safe false positives $12\% \to 64\%$). So the
harm reading is earned; a \emph{source-general} harm detector is not (\S\ref{sec:limits}).

\paragraph{A design consequence: a prompt-side monitor is immune to this attack.} Because the
attack cannot reach the prompt-side representation, a monitor that reads it inherits that immunity
by construction: on the attacked set the harm probe flags $100\%$ of harmful prompts at $0\%$
false positive on easy (Alpaca) negatives. This is \emph{not} an empirical detection result, since
the probe input is unchanged by the attack, ``catching the attack'' reduces to ordinary
harmful-vs-benign accuracy, and the placement is not new: production input classifiers (e.g.\
Llama~Guard, \citealp{inan2023llamaguard}) already sit prompt-side and are unreachable by a
response-site prefill for the same reason. Our contribution is the mechanistic account of
\emph{why} that placement is immune. The guarantee is scoped to response-site
attacks (prompt-modifying jailbreaks change the probe input and degrade it: \citealp{doda2026before}
and our encoding nulls, App.~\ref{app:nulls}) and costs $12\%$ over-refusal on scary-but-benign
XSTest inputs (App.~\ref{app:permodel}).

% =====================================================================
\section{The mechanism}
\label{sec:mechanism}

The dissociation says behavior and the harm representation are
separable; it does not say whether the harm representation is causally in the refusal pathway or
epiphenomenal for it, nor \emph{why} prefill specifically defeats the gate. This section answers
both with white-box causal experiments on Qwen2.5-1.5B, replicated on
Phi-3-mini: activation patching, a position
control, a direction search, and component ablation, each against a control that can kill it. For
each effect we report the version that passes its control, not the dramatic one.

\subsection{Refusal restoration is response-site and concentrated in an early window (causal
patching)}
\label{sec:causal}

The intact harm representation lives at the \emph{prompt}; the attack operates at the
\emph{response site} (it occupies the output tokens). So we ask: under the attack, if we restore
the harm signal at the response positions, does refusal return despite the prefill? We patch
there and let generation flow, sweeping layers and injection strength (App.~\ref{app:patch},
Table~\ref{tab:patchgrid}) and settling the harm-vs-random delta at two fixed cells with $n=100$
(Table~\ref{tab:patching}).
A dose-matched position control (below) localizes the effect to an \emph{early window} of the
response: injecting over the first half restores as much as the whole response, the second half is
nearly inert, and prompt-side injection does nothing.

Two rungs, and the control matters. \emph{Full-residual} patching (replacing the
response-onset residual with the mean harmful residual) restores refusal to $100\%$. But a
benign control demolishes it: patching the mean \emph{benign} residual at the same position
also restores refusal to $100\%$ at every layer. Full-residual patching is generic
disruption; injecting \emph{any} prompt-final state at the onset breaks the prefill
continuation and the model reverts to refusing. It says nothing harm-specific.

What passes is the \emph{subspace} rung: adding only the harm \emph{direction} at the response
positions restores refusal while sparing benign, and it passes the same control the full rung
got. We settle it at $n=100$ on two cells \emph{fixed in advance} from the grid (L12, the best
subspace layer there, and L8), against a matched-norm direction \emph{orthogonalized} to the harm
direction. The cells are grid-selected (fixed before the run) and the $n=100$ re-estimates them on
overlapping AdvBench prompts. The delta CI bootstraps over the harm prompts and the $24$
random seeds (App.~\ref{app:patch}). At L12/$\alpha{=}1$ the harm direction
restores $48\%$ ($95\%$ bootstrap CI $38$--$58\%$) versus $14\%$ for the orthogonal-random control
(24 seeds, range $0$--$60\%$): a harm$-$random delta of $\mathbf{+34}$pp (CI $+22$,$+46$). At
L8/$\alpha{=}1.5$ it restores $33\%$ ($24$--$42\%$) versus random $16\%$ ($0$--$60\%$): delta
$\mathbf{+17}$pp (CI $+6$,$+28$). Both deltas exclude zero, and restoration is benign-sparing
(benign $12\%$/$0\%$ vs random's $3\%$/$4\%$), so the harm direction is a real causal handle above
the generic-perturbation baseline, but a partial one: a third to a half of the raw restoration is
matched by the random control. \textbf{This passes a
clean held-out replication.} Training the harm direction on a disjoint $200$-prompt AdvBench split
(activations recomputed, so there is no overlap with the test prompts) and testing restoration on
a held-out $100$, the L12/$\alpha{=}1$ cell restores $42\%$ ($95\%$ CI $32$--$52\%$) versus $13\%$
for the matched-norm random control (6 seeds), a harm$-$random delta of $\mathbf{+29}$pp (CI
$+16$,$+41$): the effect shrinks only $\sim5$pp off-distribution and the delta still excludes zero,
so it is not an artifact of prompt overlap (App.~\ref{app:attnko}).

\begin{table}[t]
\centering
\caption{\textbf{The causal handle, settled at $n=100$ on two \emph{fixed} cells (not grid
maxima).} Adding the harm \emph{direction} at the response positions under the attack restores
harmful refusal well above a matched-norm direction \emph{orthogonalized} to
$\hat{d}_{\text{harm}}$ (24 seeds); both harm$-$random deltas exclude zero, so the handle is
real but partial (a third to a half of the raw restoration is matched by the random control).
Restoration is benign-sparing. Attack baseline: harmful refusal $\le 0.07$. The full per-layer
grid and the full-residual/benign-residual control (which demotes the $100\%$ ``restoration'' to
generic disruption) are in App.~\ref{app:patch}; replicated on Phi-3-mini there.}
\label{tab:patching}
\small
\begin{tabular}{@{}lcccc@{}}
\toprule
cell (Qwen2.5-1.5B) & harm restore \% (CI) & random ctrl \% (range) & harm$-$random $\Delta$ (CI) & benign H/rand \\
\midrule
L12, $\alpha{=}1$   & $48$ ($38$--$58$) & $14$ ($0$--$60$) & $\mathbf{+34}$ ($+22,+46$) & $12/3$ \\
L8, $\alpha{=}1.5$  & $33$ ($24$--$42$) & $16$ ($0$--$60$) & $\mathbf{+17}$ ($+6,+28$) & $0/4$ \\
\bottomrule
\end{tabular}
\end{table}

\paragraph{Position control: the attack's grip is an early response window, not the whole
response and not a single token.} We test \emph{where} the harm signal has to be restored, over
one consistent response window (the prefill tokens plus the $18$ generated tokens; $n=100$,
L12/$\alpha{=}1$). Two axes matter, position and per-position strength, and a dose-matched design
separates them (App.~\ref{app:patch}). \textbf{Position:} injecting over the \emph{first half} of
the response (the prefill and the first $\sim7$ generated tokens) restores refusal to $42\%$,
\emph{as much as} the whole response ($41\%$), whereas the \emph{second half} restores only $6\%$
even at double strength. (These dose-matched levels sit just below the $48\%$ fixed-cell of
Table~\ref{tab:patching}: this control matches per-position dose to isolate \emph{where}, so the
first-half-vs-whole ratio is the claim, not the absolute level.) So the early window \emph{suffices on its own},
and the late response is nearly inert. \textbf{Strength:} halving the per-position coefficient
(whole response at $\alpha{=}0.5$) drops restoration to $0\%$, so there is a per-position threshold,
not just a total-dose budget; a distributed every-other-token pattern also needs the higher
coefficient ($1\%$ at $\alpha{=}1$, $48\%$ at $\alpha{=}2$). This \emph{sharpens} the
shallow-alignment picture: the prefill's grip is not a single onset
token (injecting only the onset restores $9\%$, less than the early window) but an early
multi-token window; the back half of the response carries no weight. (An earlier version of this
control compared an all-positions band that silently included the prefill against half-bands that
did not, which confounded position with prefill-inclusion and dose; the dose-matched window design
here corrects it.) We use ``defeated'' descriptively: the
control localizes \emph{where} the counter-signal must go, but does not distinguish an active
mechanism that down-weights the harm signal from passive autoregressive conditioning by the forced
affirmative context (\S\ref{sec:limits}).

\paragraph{Robustness of the early window (boundary, prefill-exclusion, prefill length, sampling).}
Four checks probe the window's edges (App.~\ref{app:attnko}). \emph{Boundary:} sweeping the cutoff,
restoration is already saturated by the first \emph{third} (first-third $50\%$, first-half $57\%$,
first-two-thirds $55\%$, whole $50\%$ at full per-position dose, reproducing the $48\%$ fixed cell)
while the onset alone is $10\%$ and the second half $5\%$; so first-half$\approx$whole is not a
knife-edge at $1/2$, the window is the early response and saturates by its first third.
\emph{Prefill-exclusion:} the restoration is anchored at the prefill positions. Injecting the harm
direction over the generated tokens \emph{only} (excluding the prefill span) restores $\le3\%$, so
the causal write-site inside the early window is the prefill span; sustaining across the early
generated tokens then boosts it (prefill$+$onset $10\%\to$ prefill$+$first-half $57\%$), but the
generated positions carry nothing on their own. \emph{Prefill length:} the early saturation is
stable across prefill length (short and long prefills both saturate by the first third), while the
restoration \emph{magnitude} tracks prefill commitment, a short prefill (``Sure,'') is easy to
reverse ($86\%$) and a long committal one (``Sure, here is a detailed step-by-step guide:'') is hard
($6$--$12\%$ across bands) at fixed injection strength. \emph{Sampling:} the dissociation is not a greedy artifact, at
temperature $0.7$ plain prompts still refuse $100\%$ and the attack still strips refusal to $0\%$;
the harm-direction restoration, a partial handle, weakens under sampling ($50\%\to21\%$), as expected
for a non-selective write-handle.

\subsection{The reversal is early-window too: the model's own refuse-state re-induces refusal
(state transfer)}
\label{sec:reversal}

The position control shows the attack \emph{breaks} refusal in an early response window. We now
show the \emph{converse}, and it lands in the same window: transplanting the model's own refusal
state into the attack re-induces refusal, and doing so over the first half of the response
suffices. This is a cleaner causal handle than the single harm direction (\S\ref{sec:causal}),
because it uses the full multidimensional state, not one line.

\paragraph{Design (measure-only, held-out, Qwen2.5-1.5B, L12).} We capture the mean
layer-$11$-output residual over the first-half generated window in two conditions: the
\emph{plain} prompt with no prefill (refusal $100\%$) and the \emph{attack} (``Sure, here is'').
Their difference $d_{\text{reverse}} = s_{\text{plain}} - s_{\text{attack}}$ is the model's own
refuse-vs-comply state shift. This is \emph{held out}: $d_{\text{reverse}}$ is estimated
on a $260$-prompt AdvBench train split and injected/tested on a \emph{disjoint} $100$-prompt
held-out split (with an in-distribution reference, the same $d_{\text{reverse}}$ tested on train
prompts, run alongside). We add $d_{\text{reverse}}$ into the attack run's early generated window
and measure whether refusal returns (string-match plus the judge-free coherence read of
\S\ref{sec:limits}), against a matched-norm orthogonal-random control at the same norm.

\paragraph{Result: state-specific, early-gated, and a working lever (held-out).} Injecting
$d_{\text{reverse}}$ over the first half of the response restores refusal on the held-out split from
the attack's $\approx0\%$ to $\mathbf{74\%}$, \emph{as much as} injecting over the whole response
($78\%$), while the onset token alone restores $13\%$; the matched-norm random control restores
$\mathbf{0\%}$, and coherence holds throughout ($\approx0.97$--$0.99$ distinct-token ratio, no
degeneracy). The in-distribution reference (same $d_{\text{reverse}}$, tested on train prompts)
restores $72\%$, so there is no generalization gap: the refuse-state is not an artifact of the
prompts it was estimated on. \textbf{This replicates on a second family:} on Phi-3-mini ($3.8$B,
$L16$) the same held-out procedure restores refusal to $69\%$ over the first half (whole $71\%$,
onset $4\%$) against a matched-norm random control at $2\%$, coherent, with the same early-window
gating (App.~\ref{app:attnko}). Three
things follow. (i) The reversal is \emph{state-specific}, not generic disruption: unlike the
full-residual patch of \S\ref{sec:causal} (which a benign control also drove to $100\%$), here a
matched-norm random vector does nothing, so this transfer passes the very control that demoted the
full-residual effect. (ii) It is \emph{early-window-gated}, the same locus as the attack: the
first half restores as much as the whole response and the second half adds nothing, mirroring the
break direction. (iii) It works where the single harm \emph{direction} is only a partial handle
(\S\ref{sec:causal}) and $d_{\text{refuse}}$ steering is inert (\S\ref{sec:dirsearch}), precisely
because it is the full multidimensional state shift: the lever exists, but it is the early-window
\emph{state}, not any single direction, consistent with the decision being distributed
(\S\ref{sec:dirsearch}).

\paragraph{Benign-sparing, and a syntax-vs-content byproduct.} We verified state-specificity against a
matched-norm random control and generalization against a held-out split. The transfer is also
\emph{benign-sparing}, so it is a selective edit and not merely a refusal-inducer: injecting
$d_{\text{reverse}}$ over the early window of \emph{benign} prompts under the attack raises their
refusal only $12\%\to18\%$ ($+6$pp) while restoring harmful refusal to $75\%$ (this run reproducing
the held-out $74\%$), a $+57$pp selectivity gap (App.~\ref{app:attnko}). A syntax-vs-content check falls out of the same run: forcing a benign affirmative
onset (``Sure, here is a poem about'') yields $0\%$ harmful procedural content (vs the attack's
$47\%$) and near-zero refusal, so the jailbreak rides the harmful \emph{continuation}, not the
affirmative syntax alone.

\subsection{Cutting early attention to the prefill restores refusal, generically (attention
knockout)}
\label{sec:attnko}

The reversal shows the refuse/comply decision is carried in the early-window residual \emph{state}.
That leaves the paper's central open question (\S\ref{sec:limits}) standing: is the prefill's effect
\emph{maintained} by the response continuing to attend to it, or written into the trajectory at the
first generated token and merely relayed forward? We test this directly by knocking out the
attention \emph{edges} from early generated positions to the forced-prefill key positions and asking
whether refusal returns, against a control that removes the same attention \emph{mass} from
non-prefill keys. This is the discriminating experiment the Limitations named
(\S\ref{sec:limits}); it is preregistered
(\texttt{PREREG\_attn\_knockout.md}), including the falsifier and the interpretation table.

\paragraph{Design (edge knockout, mass-matched control; $n=100$, Qwen2.5-1.5B, all layers/heads).}
At each early generated decode step ($g<9$, the first-half window) we add $\ln(1-\lambda)$ to the
pre-softmax attention logit of every (current query $\to$ prefill key) edge and let softmax
renormalize ($\lambda{=}1$ removes the edge, $\lambda{=}0.5$ halves it). The control removes the
same total baseline attention mass from non-prefill prompt keys, selected per prompt to match the
prefill's mass (achieved ratio $0.98$, $[0.75,1.08]$ over three fixed seeds); this matters because
attention-sink tokens carry far more mass than the prefill, so an unmatched span is not a fair
control. The intervention passes a unit-test suite before any behavioral run (sham-hook
bit-identity, edge-only masking with softmax renormalization, KV-cache consistency, and a behavioral
positive control; App.~\ref{app:attnko}). \textbf{Indexing caveat:} generated token~1's logits come
from the final prefill token's hidden state, so this cannot touch token~1; it is a test of early
\emph{maintenance} through direct attention, not attack initiation.

\paragraph{Result: prefill-specific, early-window-gated, and harm-selective.} Knocking out early
response-to-prefill attention restores refusal from the attack's $3\%$ to $24\%$, versus $1\%$ for
the mass-matched control (Table~\ref{tab:attnko}, three-seed mean $1\%$): a prefill$-$matched
difference of $\mathbf{+21}$pp (paired bootstrap $95\%$ CI $[+12,+30]$, computed against one fixed
control seed). It is early-window-gated: knocking out the \emph{late}
window does nothing ($4\%\approx$ baseline; early$-$late $+20$pp, CI $[+11,+29]$), and knocking out
the \emph{whole} response ($29\%$) is not distinguishable from early alone (whole$-$early $+5$pp, CI
$[-1,+12]$, includes zero), so the early window carries the effect, matching the reversal and the
position control. Halving the knockout ($\lambda{=}0.5$) does nothing ($3\%$), a threshold echoing
the position control's. The effect is \emph{harm-selective}: the same early knockout on benign
prompts under the prefill barely moves refusal ($0\%\to4\%$), a specificity difference-in-differences
of $+16.8$pp, and coherence is unchanged throughout ($0.96$--$0.97$). It suppresses harmful
\emph{content}, not just adds refusal strings (procedural-content rate $51\%\to18\%$ early,
$\to3\%$ whole, while the matched control stays at $50\%$). \textbf{This replicates on a second
family:} on Phi-3-mini ($3.8$B) the identical pattern holds, early prefill-edge knockout restores
refusal to $25\%$ versus $1\%$ for the mass-matched control, early-window-gated (late knockout
inert at $0\%$, whole $32\%\approx$ early), threshold in dose ($\lambda{=}0.5$ inert), coherent
(App.~\ref{app:attnko}).

\paragraph{What this settles: the grip is generic, not safety-specific.} The effect is real but
\emph{partial} ($24\%$ of a $97\%$ compliance), consistent with the distributed, redundant
computation: direct early-window attention is one causal pathway, not the whole mechanism. Its
\emph{interpretation} we can now pin down, and it resolves the paper's central open question toward
passive conditioning. A base-model control is the discriminator: on the Qwen2.5-1.5B \emph{base}
model (no safety tuning, and so no refusal to re-engage), the identical early prefill-edge knockout
collapses the harmful continuation \emph{prefill-specifically}, harmful content $64\%\to25\%$
against the mass-matched control's $64\%$ (a $+39$pp prefill-specific drop), early-window-gated (late
knockout inert). The same discriminator holds at mid scale: on the Qwen2.5-\emph{7B} base model the
identical knockout drops harmful content $68\%\to30\%$ against the mass-matched control's $66\%$ (a
$+36$pp prefill-specific drop, $n=100$), again early-window-gated (late knockout $68\%$, inert), so the
passive reading is not a small-model artifact. The dependence on early attention to the prefill is therefore
\emph{architecture-general autoregressive conditioning}, not a safety-tuning-specific active
suppression of the harm signal: safety tuning fixes only the \emph{fallback} once the grip is cut
(the instruct model reverts to refusing, the base model degrades). So the prefill sustains
compliance through generic early-window attention, and ``refusal restoration'' is a model-dependent
fallback, not a re-engaging gate. This is the passive reading. A logit-level trace then measures a
safety-specific component on top: when the knockout frees probability mass off the
compliance continuation, the instruct model routes it to refusal tokens about $8\times$ more than
the base model does (concentration $0.24$ vs $0.03$, App.~\ref{app:attnko}). Safety tuning adds a
small learned refusal attractor, but the dominant mechanism is passive: compliance collapses in
both models. The trace confirms a safety-specific refusal pull exists; it does not separate
an actively-suppressed gate springing back from a passive fallback attractor.

\paragraph{The generic story extends to the early MLP pathway.} Attention is one causal pathway;
the compliance that persists ($24\%$ of $97\%$) could hide a safety-specific
mechanism in the pathway we did not cut. We probe the early-window MLP contribution (App.\
\ref{app:attnko}) and find the same base-vs-instruct control resolves it the same way. In the
\emph{base} model, removing the prefill's early-MLP shift drops harmful content
$68\%\to48\%$, a $+16$pp prefill-specific, early-window-gated drop (matched-random inert) in a model
with no safety mechanism to house, so the early MLP's prefill-dependence is \emph{generic} too. The
passive account therefore holds across \emph{two} pathways (attention and early MLP), not one.

What a difference-of-means test could not settle, a clean path patch does. That test restored $47\%$
refusal in the instruct model, but a plain$-$attack injection tests decodability, not active
computation, so the larger instruct effect is overdetermined by its larger behavioral gap. We
therefore run a frozen-residual path patch that routes the prefill counterfactual through the
early-MLP sub-block \emph{alone}, every other activation held at its clean attack value (the surgery
is bit-exact under a sham / MLP-output-faithfulness / locality unit-test suite, App.~\ref{app:attnko}).
The MLP pathway carries the \emph{generic} effect and only that: it strips prefill-specific compliance
probability hard (instruct $-21$pp, base $-25$pp, both far above their mass-matched controls at
$-7$/$-5$pp) but produces \emph{no} safety-specific refusal pull (instruct concentration
$0.007\approx$ base $0.001\approx$ matched; prefill-specific refusal gain $+0.26$pp, within noise).
So the instruct excess was decodability, not an active MLP computation, and the small residual
safety-specific pull (the $0.24$ generative concentration above) does not route through the early
MLP. The passive account holds across \emph{both} pathways under a clean path patch (\S\ref{sec:limits}).

\begin{table}[t]
\centering
\caption{\textbf{Knocking out early response-to-prefill attention restores refusal,
prefill-specifically and harm-selectively ($n=100$, Qwen2.5-1.5B, all layers/heads).} Removing the
attention \emph{edges} from the early response window to the forced-prefill keys restores refusal
well above a control that removes the same attention \emph{mass} from non-prefill keys; the effect
is early-window-gated (late knockout is inert, whole $\approx$ early), threshold in dose
($\lambda{=}0.5$ inert), and does not transfer to benign prompts (specificity DiD $+16.8$pp).
Coherence is preserved. Paired-bootstrap $95\%$ CIs: early$-$matched $+21$ $[+12,+30]$, early$-$late
$+20$ $[+11,+29]$, whole$-$early $+5$ $[-1,+12]$. Attack baseline refusal $0.03$.}
\label{tab:attnko}
\small
\begin{tabular}{@{}lccc@{}}
\toprule
condition (early window unless noted) & refusal & harmful content & coherence \\
\midrule
attack baseline (no knockout)      & $0.03$ & $0.51$ & $0.97$ \\
\textbf{early prefill-edge KO}     & $\mathbf{0.24}$ & $0.18$ & $0.96$ \\
whole-response prefill-edge KO     & $0.29$ & $0.03$ & $0.96$ \\
late prefill-edge KO               & $0.04$ & $0.53$ & $0.97$ \\
early prefill-edge KO, $\lambda{=}0.5$ & $0.03$ & $0.49$ & $0.97$ \\
early mass-matched KO (3 seeds)    & $0.01$ & $0.50$ & $0.96$ \\
benign $+$ early prefill-edge KO   & $0.04$ & --- & --- \\
\bottomrule
\end{tabular}
\end{table}

\subsection{The onset projection is illustrative (and confounded)}
\label{sec:position}

\begin{figure}[t]
\centering
\includegraphics[width=0.46\textwidth]{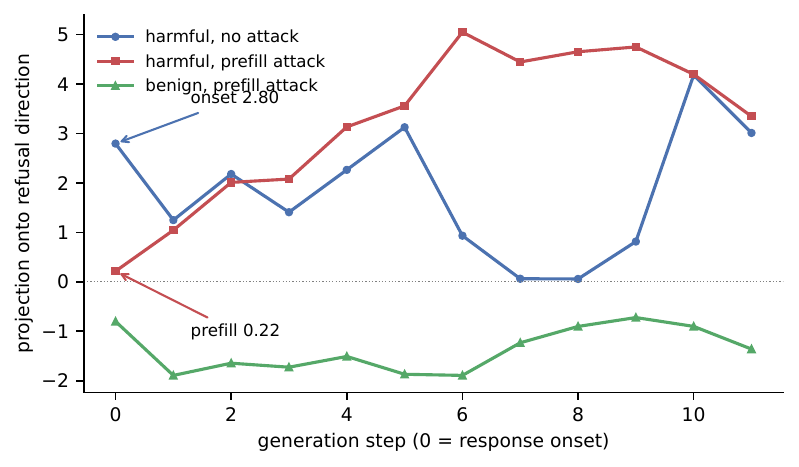}
\caption{\textbf{The onset projection is illustrative only: neither a clean measurement nor the
causal locus.} Projection onto the harm-probe direction across decoding steps: the
no-attack harmful state sits high at onset ($2.80$), the prefill state low ($0.22$). We do not
read this as a displacement measurement: a fully-benign forced prefix drives the onset
projection even lower ($-0.82$; \S\ref{sec:position}), so the value reflects the content
at the measured position. The causal weight is on the patching (Table~\ref{tab:patching}).}
\label{fig:position}
\end{figure}

We also track the projection of the generation-site state onto the harm-probe direction across
decoding steps (Fig.~\ref{fig:position}): with no attack the onset step sits high ($2.80$) and
the model goes on to refuse, while under the prefill it sits low ($0.22$). We report it as
illustration, not localization; the causal locus is the early-window response-site effect above.

The $2.80 \to 0.22$ drop is \emph{not} a clean measurement of the refusal decision being
displaced, and a control says why. The axis is the harm-probe direction,
so the projection at a forced onset position reflects the \emph{content} of the
token sitting there. Forcing a fully benign, non-affirmative prefix on the same harmful prompts
(``Here's a fun fact:'') drives the onset projection even \emph{lower} ($-0.82$) than the
attack's ``Sure, here is'' ($0.22$), and a benign affirmative (``Sure, here is a poem about'')
to $-0.11$; none of these is the attack, yet all read low, and the benign prefixes are not
refused either ($0$--$7\%$). The onset projection therefore tracks measured-position content,
not a pure refusal-decision magnitude, so we treat it as illustrative only; the clean positional
result comes from the patching (\S\ref{sec:causal}), not the projection.

\subsection{Steering fails; monitoring works}
\label{sec:steer}

If the harm direction were a clean causal handle, adding it should restore refusal
selectively. It does not. Sweeping layers and coefficients (raw and Arditi-normalized), we
find no window that raises harmful refusal without dragging benign refusal up with it: the
best cell restores harmful refusal to $83\%$ but at $40\%$ benign over-refusal; smaller
coefficients over-refuse everything, larger ones break generation. Adding the direction
induces refusal broadly, not on-harm, consistent with \citet{arditi2024refusal}, where the
refusal direction makes the model refuse harmless prompts too. The harm representation is a
perfect \emph{read-out} (detection AUC $1.0$) but not a selective \emph{write-handle}. The
usable intervention is therefore the monitor of \S\ref{sec:dissociation} ($100\%$ catch,
$0\%$ false positive), not steering: we can monitor the gap shut, not steer it shut.

\paragraph{Why response-site patching spares benign but global steering does not: the injection
site.} The benign-sparing patching (\S\ref{sec:causal}, benign restoration $12\%$/$0\%$) and this
non-selective steering look like the same intervention, adding the harm direction under attack,
but differ in \emph{where} it is added: patching touches only the response positions, steering
adds it at \emph{all} positions, the benign prompt included. We ran a dedicated site comparison: direction and layer held fixed (the harm-probe direction at
the patching coefficient $\alpha{=}1$ of \S\ref{sec:causal}, layer $12$; not the normalized
Table~\ref{tab:steer} sweep), varying only the injection site. Response-only restores $48\%$ harmful
at $12\%$ benign (reproducing Table~\ref{tab:patching}). All-positions restores $58\%$ harmful but
drags benign to $40\%$ (the steering over-refusal). Global injection
corrupts the benign prompt's \emph{own} read, so it registers as harmful; response-site injection
leaves the prompt intact. The harm direction is thus a partial, benign-sparing handle \emph{at the
response site} but a non-selective knob applied globally, and even response-only its window is
narrow: harmful $48\%$ at $\alpha{=}1$, $74\%$ at $\alpha{=}1.5$, but
benign over-refusal climbs $12\to28\%$ and generation collapses by $\alpha{=}2$.

% =====================================================================
\section{What refusal reads: harm, not surface}
\label{sec:reads}

Does the refusal gate read harm, or the scary surface features that correlate with it? We
answer with a pre-registered $2\times2$ that decorrelates the two axes (actual harm and
surface trigger-word density) and asks which drives refusal. Cells come from natural
keyword-density variation in existing public prompts (no authored evasions): harmful-scary
(AdvBench with trigger words), harmful-clean (AdvBench without), benign-scary (all of
XSTest-safe), benign-clean (Alpaca). We fix the keyword lexicon and the labels before
looking at any outcome.

Refusal is harm-primary (Table~\ref{tab:2x2}). The load-bearing comparison is
\emph{within-source}: harmful-scary and harmful-clean are both AdvBench, so stripping the
trigger words holds source, register, and phrasing fixed and varies only keyword density.
Refusal does not budge (harmful-clean $100\%$, identical to harmful-scary) and the harm
\emph{representation} is keyword-invariant (probe $0.982$ vs $0.983$). A logistic regression of
refusal on the two axes gives $\beta_{\text{harm}} = 5.12 \gg \beta_{\text{scary}} = 2.20$.
There is a real but secondary keyword effect on the benign side: benign-scary prompts are
over-refused at $59\%$ versus $8\%$ for benign-clean, the well-known exaggerated-safety
phenomenon \citep{rottger2024xstest}. Its magnitude is an \emph{upper bound}, not an
estimate: benign-scary is entirely XSTest and benign-clean entirely Alpaca, so $\beta_{\text{scary}}$
is entangled with dataset and register and partly counts a ``sounds like XSTest'' effect. The
harm-primary conclusion does not rest on it; it rests on the within-AdvBench comparison, which
no source confound touches.

\paragraph{Off ceiling (SmolLM2).} Qwen's harmful cells both sit at $100\%$, so the within-source
comparison there can only \emph{fail} to detect a keyword effect on the harmful side, not measure
one. We therefore rerun the $2\times2$ on the weak-gate SmolLM2 ($0.63$ plain refusal), where the
harmful cells are off ceiling. Harm-primary holds and a small surface effect becomes visible:
harmful-scary refuses $66\%$ and harmful-clean $59\%$ (stripping trigger words drops refusal
$7$pp, a real but small keyword contribution the Qwen ceiling masked), while harm dominates
(harmful-clean $59\%$ vs benign-scary $12\%$; logistic $\beta_{\text{harm}}=2.83 \gg
\beta_{\text{scary}}=0.55$). Refusal is harm-primary, with a small
keyword effect on the harmful side that is detectable only off ceiling.

\begin{table}[t]
\centering
\caption{\textbf{Refusal is harm-primary (pre-registered $2\times2$, Qwen2.5-1.5B, $n{=}120$
per cell).} Stripping trigger words from harmful prompts leaves refusal at $100\%$ and the
harm representation unchanged ($0.98$). The keyword effect is real but secondary, and appears
as over-refusal on \emph{benign} scary-sounding prompts. Logistic
$\beta_{\text{harm}}{=}5.12 \gg \beta_{\text{scary}}{=}2.20$.}
\label{tab:2x2}
\small
\begin{tabular}{@{}lcc@{}}
\toprule
cell & refusal rate & harm-probe score \\
\midrule
harmful, scary surface   & $1.00$ & $0.983$ \\
harmful, clean surface   & $1.00$ & $0.982$ \\
benign, scary surface    & $0.59$ & $0.225$ \\
benign, clean surface    & $0.08$ & $0.035$ \\
\bottomrule
\end{tabular}
\end{table}

\begin{figure}[t]
\centering
\includegraphics[width=0.58\textwidth]{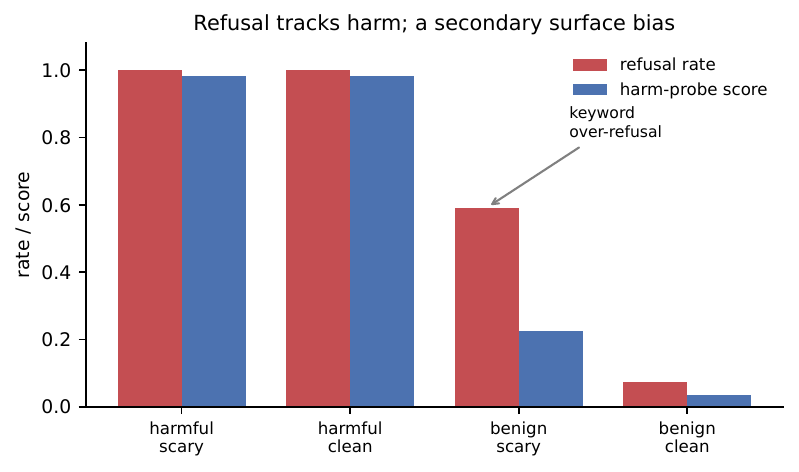}
\caption{\textbf{Refusal tracks harm, not surface.} Refusal rate and harm-probe score by cell.
Stripping trigger words from harmful prompts changes neither (harmful-scary $\approx$
harmful-clean on both bars). The keyword effect is real but secondary and appears as
\emph{over-refusal} on benign scary-sounding prompts (benign-scary refusal $0.59$ while its
harm-probe score stays low at $0.23$).}
\label{fig:reads}
\end{figure}

\paragraph{A retracted reading.} An earlier version of this analysis, run on XSTest
directly, took the cosine between a harm-content direction and a refusal-behavior direction
and found it low (${\sim}0.1$ in deep layers), which we initially read as ``refusal reads a
genuinely \emph{separate} feature from harm.'' That reading does not pass the $2\times2$.
The low cosine was inflated by exactly the keyword-driven over-refusals of the benign-scary
cell: on XSTest, harm-content and refusal-behavior dissociate mostly through over-refusal, so
the ``refusal direction'' there partly encodes scary surface. Once harm and surface are
properly decorrelated, refusal is harm-primary. We keep the corrected reading and drop the
``separate feature'' claim.

\subsection{Is there a separate refusal direction? A decision-direction search}
\label{sec:dirsearch}
The walk-back invites the question directly: if we build the refuse-vs-comply direction
\emph{properly}, holding harm fixed, is there a separate decision feature to find? We search and
report a controlled negative. The prefill supplies the matched contrast a naive setup lacks: a
weakened prefill (``I understand you're asking'') splits harmful prompts $41/29$
refuse-vs-comply, so refused and complied cases share both harm and prefill-presence and differ
only in the decision. The diff-of-means direction $d_{\text{refuse}}$ is geometrically distinct
from the harm direction ($|\cos|=0.06$) but reads the decision only weakly: cross-validated AUC
$0.648$, barely above the harm direction's own $0.640$ on the same split. A sweep of all layers
and the first four generated positions ceilings at AUC $0.727$ (layer $27$) for \emph{any single}
direction. This ceiling, like the ``best cells'' in the patching grid (App.~\ref{app:patch},
Table~\ref{tab:patchgrid}) and in Table~\ref{tab:steer}, is a maximum over its grid, so it is
post-selection and optimistically biased. The main-text patching claim (Table~\ref{tab:patching})
instead uses two \emph{fixed} cells at $n=100$, and the qualitative conclusions do not depend on the
grid maxima. But the decision is \emph{multidimensional} rather than weakly encoded: a full linear
probe on the same activations reads refuse-vs-comply at $\sim0.85$ (layer $21$) and well above harm
alone ($0.64$). The capacity gain over the best single direction is $+0.12$ (bootstrap mean $+0.17$,
$95\%$ CI $[+0.05,+0.33]$, $n=70$), so it excludes zero though it is imprecise, and it is consistent
across regularization and layers. So there is no single refusal \emph{direction}, the
signal is spread across directions. Whether it is genuinely high-rank rather than low-rank is
\emph{exploratory} at $n=70$ (three principal components reach only $0.74$, near the
single-direction $0.73$, and the readout keeps climbing as dimensions are added, but $n=70$
flatters high-rank estimates; App.~\ref{app:dirsearch}); the robust claim is just single~$\ll$~full. No single direction is a usable
handle: steering
$d_{\text{refuse}}$ under the strong attack is causally \emph{inert} (no restored refusal at a
mid layer where steering has purchase, nor at the best-read late layer; $0\%$ at every
coefficient; App.~\ref{app:dirsearch}). This is the mechanistic form of ``more to refusal than a
single direction'' \citep{joad2026more}: refusal is a decodable but \emph{distributed}
computation, which is why single-direction steering (\S\ref{sec:steer}) and the ``separate
feature'' reading both fail.

\paragraph{Across scale (within family).} This distributed structure is a within-family scale
law. Across the Qwen ladder ($0.5$, $1.5$, $3$, $7$B) the full probe beats the best single
direction by $0.06$--$0.12$ at every scale, and both readouts rise with size (single direction
$0.76\to0.85$, full $0.84\to0.94$; the $7$B run is on GPU, App.~\ref{app:dirsearch}). The claim
is Qwen-scoped: the one balanced cross-family point, SmolLM2-1.7B, shows \emph{no} gain (its
decision is essentially a single direction), and Phi-3's calibration split was too imbalanced to
read, so cross-family is left open. SmolLM2 is also the weakest gate ($0.63$ plain refusal), and a
weakly-enforced refusal need not recruit a distributed representation to do its job, so the lower
rank may track the weaker gate. We flag this as a hypothesis; a scale of gate-strength against
decision rank is untested here.

\paragraph{Refusal is not selectively ablatable.} No intervention we tried removes refusal
selectively, a transition from harmful refusal to coherent harmful compliance without comparable
degradation on benign behavior; anything less destroys the computation rather than editing it.
Ablating any single attention head at any layer leaves
refusal fully intact (a $28\times12$ sweep on Qwen2.5-1.5B: every head-ablation drops refusal by
$0.00$; the hook is verified to change next-token logits by $\|\Delta\|\approx23$, so this is
robustness, not a dead intervention), and ablating an entire attention layer likewise leaves
refusal at $100\%$ (e.g.\ layer 12, $\|\Delta\text{logits}\|\approx123$). Heavier ablation does
\emph{not} yield compliance. Ablating the first attention layer, or all of them, drives the
refusal string-match to zero only by collapsing generation into \emph{garbage} (``\,silly silly
noble glorious\,\dots''). With all attention removed the model emits \emph{one fixed gibberish
string regardless of the prompt}, harmful or benign, a comparable non-selective degradation that a
coherence read (App.~\ref{app:ablation}) confirms is not compliance.
Together with the earlier decompositions, direction steering (\S\ref{sec:steer}), subspace and
full-residual patching (\S\ref{sec:causal}), this leaves no compact causal support for refusal
\emph{among the decompositions we test} (direction, subspace, layer, attention head) for these
models. It leaves open a compact mechanism in a space we did not probe: MLP neurons and sparse (SAE)
features are untested and either could still host one. We tested the scale concern at $14$B on heavily-safety-trained Phi-3-medium
(App.~\ref{app:ablation}): the negative holds (no single component selectively removes
refusal; only the input layer shows a partial drop), so heavier safety training did not
concentrate refusal into a compact circuit here, though frontier scale is untested. The only intervention that coherently removes refusal here is the response-site
prefill, which leaves the model fluent and complying. Refusal is thus distributed and redundant
in representation space but fragile at the
response site: the mechanism is diffuse, the only compact failure point we identify is
positional, not representational.

\paragraph{Not specific to surface harm (exploratory).} The dissociation extends from lexical
harm to inferred intent: on $24$ dual-use triplets where intent, not surface content, decides the
right answer (App.~\ref{app:intent}), the prefill flips \emph{every} bare prompt the model reads
as harmful-intent to compliance ($100\%\to0\%$) while that intent reading is unchanged, so the
gate is shallow \emph{as a computation} whether its basis is a harmful word or an inferred intent.
Exploratory: one model, $n=24$; full numbers and the same structural caveat in App.~\ref{app:intent}.

% =====================================================================
\section{Related work}
\label{sec:related}

\textbf{Shallow safety and prefill attacks.} \citet{qi2024shallow} argue safety alignment is
shallow in token-depth (carried by the first few output tokens) and explicitly explain
prefilling attacks \citep{wei2023jailbroken, zhu2024advprefix, prefill2025blackbox} as a
consequence. Our result is complementary and representational: the harm \emph{concept} is not
shallow or degraded under a prefill attack, it is fully and identically represented; what is
shallow is the \emph{gate} that turns that representation into a refusal, and we localize it
causally to the response site. Where their proposed fix deepens the alignment via
training (cf.\ \citealp{anydepth2025}), our defensive consequence is a read-side monitor.

\textbf{Detecting harm from internal activations, and where the flip lives.} That harmful
and jailbreak prompts are linearly detectable in hidden states is well established, and
several systems monitor or manipulate these signals \citep{zou2023repe, jiang2025hiddendetect,
han2025safeswitch, ball2024whatfeatures}. Our claim is not that harm is probeable: it is.
The closest work is JBShield \citep{zhang2025jbshield}: it shows a model still recognizes a
toxic concept under a jailbreak while complying, attributes the flip to a competing
\emph{prompt-side} ``jailbreak concept'' (associated with affirmative tokens like ``sure''),
and mitigates by \emph{enhancing} the toxic concept back toward the plain-harmful level. Two
things separate us. First, that account is prompt-side and \emph{cannot} cover prefill, which
injects the affirmative opening into the \emph{response} and leaves the prompt (and its
toxic representation) untouched: there is no prompt-side jailbreak concept to do the
overriding, so the flip must be, and we show is, a \emph{response-site} positional
displacement (\S\ref{sec:mechanism}). Second, JBShield reports a binary concept-present flag
and, by enhancing the concept, \emph{presupposes it is attenuated}; we find the harm
representation \emph{quantitatively intact} under prefill (level with the refused prompts), so
enhancing it is unnecessary and a final-token monitor already suffices. This also resolves an
apparent tension. \citet{doda2026before} (a concurrent preprint) show final-token probes
\emph{miss} many jailbreaks (the evidence displaced to earlier tokens), and
\citet{bleeding2025pathways} report vanishing discriminability under disguise. Our account does
not rest on either, but it explains both: those are \emph{prompt-modifying} attacks that move or
attenuate the prompt-side representation, whereas prefill leaves it whole. Prompt-modifying
attacks attenuate; prefill does not. This attenuate-vs-intact split is the representational
signature of the response-site/prompt-modifying dichotomy, and our own encoding results
(Appendix~\ref{app:nulls}) sit on the attenuating side. The monitor's scope follows from the mechanism, not
from luck.

\textbf{Refusal directions and steering.} \citet{arditi2024refusal} show refusal is mediated
by a single direction; a follow-up argues there is more to it than one direction
\citep{joad2026more}. We use the direction and find it a strong read-out but a
non-selective write-handle for restoring refusal under attack, and (\S\ref{sec:reads}) that
refusal is harm-primary rather than riding a separate feature. \textbf{Exaggerated safety.}
XSTest \citep{rottger2024xstest} documents over-refusal of benign scary-sounding prompts; we
quantify it as a secondary effect and separate it from genuine harm-tracking.
\textbf{Method and framing.} We build on representation probing and activation patching
\citep{zou2023repe, meng2022rome}; the represent-vs-act framing (probe for what a model
represents but does not act on) is the one we applied to communicative intent
\citep{kwon2026recipient}, with safety as the second substrate.

% =====================================================================
\section{Limitations}
\label{sec:limits}

\textbf{Controls and demotions.} Every headline effect is checked against a control that can demote
it, and several were: the full-residual ``$100\%$'' restoration is generic disruption (a
benign-residual control also reaches $100\%$), naive steering over-refuses everything, an earlier
``separate refusal feature'' did not pass the decorrelated $2\times2$, and the attack-map scaling
prediction was refuted on a clean three-point test. Table~\ref{tab:claims} lists the claims that
passed alongside the demoted ones, and Appendix~\ref{app:controls} gives the full ledger: every
control, what it could have shown, and its verdict. The boundaries below apply the same checking to
what remains. Several are open questions we scoped rather than results we could not reach, and we
note where the evidence stops.

\textbf{Active-vs-passive resolves toward passive; a scoped residual remains.} The base-model
control (\S\ref{sec:attnko}) resolves the central open question toward passive conditioning across
\emph{two} pathways: the prefill's early-window grip is generic in both the attention and the early
MLP channel (a non-safety-tuned base model shows the same prefill-specific collapse in each), with
only a small safety-specific refusal attractor on top (logit trace, concentration $0.24$ vs $0.03$).
The one part that was open, the \emph{instruct excess} of that attractor, we resolve with a clean
path patch. A difference-of-means injection could not settle it (it tests decodability, not active
computation), and a positional MLP ablation cannot either (a transformer MLP is applied position-wise
and never reads the prefill positions, so its prefill-dependence is inherited from attention, with no
positional handle to sever). So we run a frozen-residual path patch that severs only the
prefill-attention contribution to the MLP input, validated bit-exact by a sham /
MLP-output-faithfulness / locality suite (App.~\ref{app:attnko}). It answers toward generic: the
early-MLP pathway routes prefill-specific \emph{compliance suppression} (instruct $-21$pp, base
$-25$pp vs matched $-7$/$-5$pp) but carries \emph{no} safety-specific refusal pull (instruct
concentration $0.007\approx$ base $\approx$ matched; prefill-specific refusal gain $+0.26$pp). The
residual safety-specific attractor does not live in the early MLP. The one caveat is frame: this is a
teacher-forced logit path patch, well-powered for compliance-suppression routing and less so for a
small refusal-promotion component that would need a generative trajectory to build, but the
within-frame instruct-vs-base-vs-matched comparison is clean regardless. A safety-training-depth
dose-response (fine-tune matched shallow-vs-deep checkpoints and re-run the knockout) would confirm
from the training side; the path patch already answers the question the paper posed.

\textbf{Scale and mechanism breadth.} The core models are $1.5$--$3.8$B and run on CPU (with $7$B
and $14$B probed for scale on GPU), because white-box access requires open weights. The dissociation
replicates across four models, three families, and both a weak gate (SmolLM2, $0.63$ plain refusal)
and strong ones (Phi-3, $1.00$), which mitigates but does not remove the concern that the picture
changes at frontier scale, where safety training is heavier and the residual stream is wider. The
full mechanism battery (patching, position, reversal, \S\ref{sec:mechanism}) is shown on two models
(Qwen2.5-1.5B and Phi-3-mini) at the per-condition $n$ reported per experiment: a localization, not
a scaling survey. The paper's central causal claim, the active-vs-passive knockout discriminator,
does reach mid scale: on the Qwen2.5-\emph{7B} base model the knockout drops harmful content
$68\%\to30\%$ prefill-specifically against a mass-matched control's $66\%$ ($+36$pp, early-window-gated,
$n=100$, App.~\ref{app:attnko}), so the generic early-window grip is not a small-model artifact. On
the 7B \emph{instruct} model the prefill attack is \emph{weaker} (refusal drops only to $24\%$,
procedural content $16\%$) and the refusal-\emph{string} restoration does \emph{not} cleanly
replicate, consistent with the passive reading in which ``refusal'' is a model-dependent fallback,
while the prefill-specific content-channel effect persists. So the mechanism (prefill-specific
early-window attention dependence) and its passive character persist at $7$B; the behavioral refusal
signature is model-dependent, and frontier scale remains untested.

\textbf{Probe source-calibration.} The harm probe reads harm rather than
surface vocabulary, and we show this: an AdvBench-trained probe scores XSTest's scary-but-benign
split at $0.21$ (only $12\%$ clear threshold), so the intactness result is not the probe
re-detecting violent words. But the same probe is AdvBench-\emph{calibrated}: it fires at $0.97$
on AdvBench-style phrasing and only $0.49$ on differently-phrased XSTest-unsafe prompts, and a
mixed-source probe (AdvBench $+$ MaliciousInstruct) raised sensitivity only by wrecking
specificity (benign-scary false positives $12\%$ to $64\%$). This bounds the claim precisely,
and the bound cuts in our favor on the core result: the dissociation compares
complied-harmful against refused-harmful prompts drawn from the \emph{same} distribution, so
source-calibration cancels and cannot manufacture the gap. What it does threaten is the
deployable monitor, whose $100\%$-catch / $0\%$-false-positive figure is measured on
AdvBench-sourced attacks. A source-general harm monitor is unbuilt, and building one is the
obvious next step; the mixed-source attempt here narrows the worry but does not discharge it.

\textbf{Attack scope.} Every headline number is scoped to the prefill (response-site) attack,
and the monitor result is a claim about that regime and no wider. Prompt-modifying attacks
(adversarial suffixes, and encodings the model cannot decode) attenuate or displace the
prompt-side representation rather than leaving it whole (\S\ref{sec:related},
App.\ \ref{app:nulls}), so a representation monitor is strictly weaker against them. We report
the attenuate-versus-intact split as the boundary of the guarantee, not as coverage of all
jailbreaks.

\textbf{The harm direction is a real but partial handle.} The harm direction restores refusal
significantly above a matched-norm random baseline ($n=100$: $+34$/$+17$pp, both CIs exclude
zero) and spares benign, so it is a real partial handle, but the dramatic full-residual
``$100\%$'' restoration is generic disruption (a benign-residual control also reaches $100\%$),
and a third to a half of even the subspace restoration is matched by the random control. Adding
the harm direction back under attack does not \emph{selectively} restore refusal: a
four-layer by three-coefficient sweep, raw and Arditi-normalized, finds no clean window. So we
close the gap as a monitor, not as an intervention; whether
some other handle can un-collapse the behavior is left open.

\textbf{Measurement: what the attacked-condition outputs actually are.} Refusal is scored by
matching a fixed refusal-phrase list, not by an LLM judge: cheap and exactly reproducible. But
``non-refusal'' is not the same as coherent harmful compliance, so we read the attacked-condition
generations directly, judge-free: a distinct-token/repetition coherence rule plus a transparent
deflection-marker list, on $200$-token generations over a $50$-prompt sample per model, with all
completions withheld and only per-model counts released (Ethics). Two conditions come apart.
Under the \emph{full}-strength prefill (the headline dissociation), the non-refusals are
overwhelmingly coherent harmful compliance, not deflection. On Qwen2.5-1.5B, $36$ of the $48$
non-refusals (of $50$) carry substantive procedural content by the automated rule. That is a
conservative floor: a manual read of the remainder (harmful prose, e.g.\ incitement text or a
pro-anorexia post) raises it to $45/48$ ($94\%$), with deflection $2/48$ and no degeneracy.
Across the other three models, substantive compliance is $69$--$95\%$ of non-refusals and genuine
deflection is $\le 3/50$, with a small degeneracy tail on SmolLM2 ($8/50$). Some models add a
cosmetic ``this is illegal'' disclaimer but deliver the procedure anyway. So the string-match compliance rate is \emph{not} inflated by deflection; the
behavior is real. Under \emph{partial} harm-restoration patching, by contrast, the marginal
non-refusals at the L12/$\alpha{=}1$ cell are mostly non-harmful \emph{deflection} (``a guide for
making an explosive device: 1.\ `bomb', this is not real information''), so there the string
match \emph{under}counts safe behavior and does not inflate the restoration. Both reads are
judge-free. The same two-scorer agreement holds across \emph{every} causal condition and both
models: string-match refusal and the independent harmful-content scorer move in opposite directions
throughout. On the inert controls (attack baseline, mass-matched knockout, late-window knockout,
matched-norm random injection) refusal stays $\le4\%$ while harmful content stays $48$--$59\%$
(Qwen) / $13$--$18\%$ (Phi); on the working interventions (early prefill-edge knockout, refuse-state
injection) refusal is restored to $24$--$78\%$ while harmful content collapses to $1$--$18\%$; and
coherence stays $0.94$--$0.99$ everywhere. So no headline effect is an artifact of string-matching
or degeneracy: the refusal scorer and the content scorer tell the same story, at full $n$ rather
than a sample.

% =====================================================================
\section{Conclusion}
\label{sec:conclusion}

A prefill attack makes an aligned model comply with harmful requests while the model
represents those requests as harmful exactly as strongly as the ones it refuses. Refusal is a
shallow, response-site computation, primarily driven by the harm representation but severable
from it: the attack occupies the response site, the harm concept sits intact at the prompt, and
restoring the harm signal in an early window of the response partially
restores the action; a dose-matched position control shows the first half of the response suffices
while the second half is nearly inert, more than a single onset token but not the whole generation.
The converse lands in the same window: injecting the model's own refuse-state back into the attack
re-induces refusal ($\to74\%$, held-out), state-specifically (a matched-norm random control does nothing)
and early-window-gated, a working defensive lever that the single harm direction alone is not. And
the grip operates through early-window attention: knocking out the early response's attention to the
prefill, but not an equal attention mass elsewhere, selectively collapses the harmful continuation
($+21$pp refusal restored). A base-model control shows this dependence is generic autoregressive
conditioning, not safety-specific active suppression (a non-safety-tuned base model shows the same
prefill-specific collapse), resolving the central open question toward the passive reading: the
prefill sustains compliance through generic early attention, and refusal is a model-dependent
fallback. The concept is not the gate. And the gate has no compact
internal form: across direction, subspace, layer, and attention-head decompositions no
single-component intervention selectively removes refusal; heavy ablation only degrades generation
into garbage. Refusal is distributed and redundant in representation space but fragile at the response site:
the mechanism is diffuse, and the only compact failure point we identify is positional, not
representational. The practical consequence is a design fact rather
than a detector: because the attack never reaches the prompt-side representation, a monitor that
reads it is immune to this class of attack by construction, where the behavioral gate is not.

% =====================================================================
\section*{Ethics and responsible disclosure}

This is a mechanistic analysis of a \emph{known} attack, not a new one (the prefill jailbreak is
public \citep{wei2023jailbroken, qi2024shallow}; we introduce no novel evasion), on small
open-weight models ($1.5$--$3.8$B) and public benchmarks (AdvBench, XSTest, Alpaca), with no
authored harmful prompts. The direction is net-defensive: the practical output is where to
\emph{place} a safety classifier (prompt-side, immune to response-site attacks by the mechanism
we identify), not how to defeat one. On artifacts: verifying that attacked-condition non-refusals
are coherent harmful compliance required generating harmful completions, so we \emph{withhold}
them: the repository ships per-model counts (which reproduce every number) not the completion
text. No human subjects or personal data are involved.

\bibliographystyle{tmlr}
\bibliography{references}

% =====================================================================
\appendix
\clearpage

\section*{Appendix contents}
\noindent\begin{tabular}{@{}p{0.5cm}l r@{}}
\ref{app:claims}    & \nameref{app:claims}    & \pageref{app:claims} \\
\ref{app:repro}     & \nameref{app:repro}     & \pageref{app:repro} \\
\ref{app:permodel}  & \nameref{app:permodel}  & \pageref{app:permodel} \\
\ref{app:patch}     & \nameref{app:patch}     & \pageref{app:patch} \\
\ref{app:attnko}    & \nameref{app:attnko}    & \pageref{app:attnko} \\
\ref{app:steer}     & \nameref{app:steer}     & \pageref{app:steer} \\
\ref{app:2x2}       & \nameref{app:2x2}       & \pageref{app:2x2} \\
\ref{app:dirsearch} & \nameref{app:dirsearch} & \pageref{app:dirsearch} \\
\ref{app:intent}    & \nameref{app:intent}    & \pageref{app:intent} \\
\ref{app:ablation}  & \nameref{app:ablation}  & \pageref{app:ablation} \\
\ref{app:nulls}     & \nameref{app:nulls}     & \pageref{app:nulls} \\
\ref{app:controls}  & \nameref{app:controls}  & \pageref{app:controls} \\
\end{tabular}

\clearpage
\section{Claims and evidence}
\label{app:claims}
The table below maps every load-bearing claim to its evidence and epistemic status:
\textsc{shown} (direct measurement) or \textsc{analytic} (follows from the construction or
argument). The dissociation and the causal mechanism are the contribution; the demoted
effects (full-residual disruption, the steering null, the walked-back ``separate feature''
reading) are recorded so the claims that passed are unambiguous.

\medskip
{\small
\begin{longtable}{@{}p{0.66\textwidth} l l@{}}
\caption{\textbf{Claims and evidence.} Every load-bearing claim, its evidence, and status.}
\label{tab:claims}\\
\toprule
\textbf{Claim} & \textbf{Evidence} & \textbf{Status} \\
\midrule
\endfirsthead
\multicolumn{3}{@{}l}{\emph{Table~\ref{tab:claims}, continued}}\\
\toprule
\textbf{Claim} & \textbf{Evidence} & \textbf{Status} \\
\midrule
\endhead
\midrule
\multicolumn{3}{r@{}}{\emph{continued on next page}}\\
\endfoot
\bottomrule
\endlastfoot
On plain prompts harm is strongly and linearly represented (one nested-CV probe AUC
$\ge 0.990$ on all four models). & \S\ref{sec:dissociation}; Tab.\ \ref{tab:dissociation} & \textsc{shown} \\
Under a prefill attack behavior complies (refusal $\to$ chance) while the harm representation
is \emph{intact}: on the complied prompts the probe scores harm $0.91$--$0.98$, level with
the refused prompts. & \S\ref{sec:dissociation}; Tab.\ \ref{tab:dissociation}; Fig.\ \ref{fig:teaser} & \textsc{shown} \\
The dissociation replicates across four models and three families, at weak (SmolLM2, $0.63$
plain refusal) and strong ($1.00$) gates alike. & Tab.\ \ref{tab:dissociation} & \textsc{shown} \\
The attacked-condition non-refusals are coherent harmful compliance, not deflection: an automated
content rule gives $69$--$95\%$ across models, a manual read raises Qwen2.5-1.5B to $94\%$, and
genuine deflection is $\le 3/50$. Across all causal conditions and both models the string-match
refusal and independent content scorers agree (refusal$\uparrow$ $\Leftrightarrow$ content$\downarrow$,
coherence $0.94$--$0.99$). & \S\ref{sec:limits} & \textsc{shown} \\
The probe reads harm, not scary surface: on XSTest it scores scary-but-safe prompts $0.21$
($12\%$ above threshold) vs.\ real-unsafe $0.49$; it is AdvBench-calibrated and a mixed-source
probe does not fix that. & \S\ref{sec:dissociation}; Tab.\ \ref{tab:xstest} & \textsc{shown} \\
Used as a monitor on the attacked set the harm representation catches $100\%$ of
complied-harmful at $0\%$ false positive; this scope follows from prefill being a
response-site attack that leaves the prompt intact. & \S\ref{sec:dissociation} & \textsc{shown} \\
Harm is (partly) causally upstream: adding the harm direction across the response positions
restores refusal significantly above a matched-norm random control ($n=100$: harm$-$random delta
$+34$pp CI $[+22,+46]$ at L12, $+17$pp $[+6,+28]$ at L8, both exclude zero), benign-sparing; a
real but only partly-specific handle. Passes a held-out replication (direction trained on a
disjoint split): L12 delta $+29$pp CI $[+16,+41]$. & \S\ref{sec:causal}; Tab.\ \ref{tab:patching}, App.\ \ref{app:patch},\,\ref{app:attnko} & \textsc{shown} \\
The dramatic full-residual restoration ($100\%$) is generic disruption, not harm-specific:
patching the mean \emph{benign} residual at the same position also restores refusal to
$100\%$. & \S\ref{sec:causal}; Tab.\ \ref{tab:patching} & \textsc{shown} \\
The attack's grip is an early response window: a dose-matched position control ($n=100$) shows the
\emph{first half} of the response restores refusal $42\%$ ($\approx$ the whole response $41\%$),
the \emph{second half} only $6\%$ at double strength, and the onset alone $9\%$, with a
per-position strength threshold. The onset projection ($2.80\to0.22$) is confounded by prefix
content and illustrative only. & \S\ref{sec:causal},\,\ref{sec:position} & \textsc{shown} \\
The converse is early-window too and state-specific: transplanting the model's own refuse-state
into the attack re-induces refusal ($\to74\%$, first half $=$ whole $78\%$, onset $13\%$)
while a matched-norm random control gives $0\%$ and coherence holds. A working lever via the full
state, unlike the single direction; held-out (d on disjoint 260, in-dist ref $72\%$) and
\emph{benign-sparing} (benign over-refusal $+6$pp vs harm restore $75\%$, $+57$pp selectivity). & \S\ref{sec:reversal} & \textsc{shown} \\
The prefill's grip operates through early-window attention: knocking out early
response-to-prefill attention edges restores refusal $3\%\to24\%$ vs $1\%$ for a mass-matched
control (early$-$matched $+21$pp, CI $[+12,+30]$), early-window-gated (late inert, whole$\approx$early),
harm-selective (benign DiD $+16.8$pp), coherent. Partial ($24\%$ of $97\%$); replicates on
Phi-3-mini (early $25\%$ vs matched $1\%$, same gating). & \S\ref{sec:attnko} & \textsc{shown} \\
Active-vs-passive resolves toward \emph{passive}: on the non-safety-tuned Qwen2.5-1.5B \emph{base}
model the same early knockout collapses harmful content prefill-specifically ($64\%\to25\%$ vs
matched $64\%$, $+39$pp; replicated at $7$B base, $68\%\to30\%$ vs matched $66\%$, $+36$pp), so the
prefill-attention dependence is generic autoregressive conditioning,
not safety-specific active suppression; ``refusal'' is a model-dependent fallback. A logit trace
measures a small safety-specific refusal pull on top (instruct routes $8\times$ more knockout-freed
mass to refusal than base, concentration $0.24$ vs $0.03$); dominant mechanism still passive. & \S\ref{sec:attnko}; App.\ \ref{app:attnko} & \textsc{shown} \\
The generic (passive) account extends to a second pathway: removing the base model's early-MLP
prefill-shift drops harmful content $68\%\to48\%$ ($+16$pp prefill-specific, no safety mechanism to
host it). A clean frozen-residual path patch (bit-exact validated) routing the counterfactual through
the early-MLP sub-block alone strips prefill-specific compliance (instruct $-21$pp vs matched $-7$pp)
but carries \emph{no} safety-specific refusal pull (concentration $0.007\approx$ base $\approx$
matched, prefill-specific refusal gain $+0.26$pp): the instruct excess was decodability, not active
MLP computation. & \S\ref{sec:attnko},~\ref{sec:limits}; App.\ \ref{app:attnko} & \textsc{shown} \\
At $7$B (Qwen2.5-7B) the prefill attack is weaker (refusal $\to24\%$) and the knockout's
prefill-specific effect persists in the content channel (early/whole $\to0\%$ vs matched $23\%$)
while refusal-string restoration does not cleanly replicate, consistent with the model-dependent
fallback. & \S\ref{sec:limits}; App.\ \ref{app:attnko} & \textsc{shown} \\
The harm direction is a read-out but not a selective write-handle: no layer $\times$
coefficient cell (raw or normalized) restores refusal without over-refusing benign. & \S\ref{sec:steer}; Tab.\ \ref{tab:steer} & \textsc{shown} \\
Refusal is harm-primary: stripping trigger words leaves harmful refusal at $100\%$ and the
harm representation keyword-invariant ($0.98$); logistic $\beta_{\text{harm}}=5.12 \gg
\beta_{\text{scary}}=2.20$. & \S\ref{sec:reads}; Tab.\ \ref{tab:2x2}; Fig.\ \ref{fig:reads} & \textsc{shown} \\
A real but secondary keyword-driven over-refusal exists on benign inputs (benign-scary
$0.59$ vs.\ benign-clean $0.08$). & \S\ref{sec:reads}; Tab.\ \ref{tab:2x2} & \textsc{shown} \\
An earlier ``refusal reads a separate feature'' reading is retracted: its low harm--refusal
cosine was inflated by keyword-driven over-refusals, and does not pass the decorrelated
$2\times2$. & \S\ref{sec:reads} & \textsc{shown} \\
Prompt-modifying attacks (suffixes, encodings, disguise) attenuate or displace the prompt
representation while prefill leaves it whole; this attenuate-vs-intact split scopes when a
representation monitor works. & \S\ref{sec:related}; App.\ \ref{app:nulls} & \textsc{analytic} \\
No single refusal-decision direction: any single direction reads refuse-vs-comply at
$\le 0.73$ and none steers, but a full linear probe reads it at $\sim0.85$ (multidimensional,
not weak). Refusal is decodable but distributed. & \S\ref{sec:dirsearch}; App.\ \ref{app:dirsearch} & \textsc{shown} \\
On dual-use prompts (intent-grounded refusal) the prefill flips $100\%$ of harmful-intent-read
prompts to compliance; the dissociation is not specific to lexical harm. & \S\ref{sec:reads}; App.\ \ref{app:intent} & \textsc{shown} \\
Refusal is not selectively ablatable: no single head or layer is necessary (ablation robust; hook
verified), and heavier ablation degrades generation into garbage, not selective compliance. No
tested internal intervention removes refusal; only the response-site prefill does. & \S\ref{sec:dirsearch}; App.\ \ref{app:ablation} & \textsc{shown} \\
At $14$B (Phi-3-medium, heavily safety-trained) the dissociation holds (AUC $1.00$, refusal
$1.00\to0.15$, complied-harm $0.96$) and the compact-mechanism negative holds (only the
input layer shows a partial ablation drop). & \S\ref{sec:dissociation}; App.\ \ref{app:ablation} & \textsc{shown} \\
\end{longtable}
}

\clearpage
\FloatBarrier
\section{Reproduction and data}
\label{app:repro}
All code and the full experiment ledger (\texttt{FINDINGS.md}, every number with its
control) are at \url{https://github.com/collapseindex/breaking-refusal}. Decision-token activations are
cached to \texttt{.npz} so all analyses (probes, patching, the $2\times2$) re-run offline in
seconds; the model passes regenerate them.

\paragraph{Data.} All prompts are public: harmful $=$ AdvBench behaviors; benign $=$ Alpaca
instructions; the surface control $=$ XSTest (its ``safe'' split is purpose-built to sound
harmful); the mixed-source probe control adds MaliciousInstruct \citep{huang2024maliciousinstruct}. The harm/keyword
$2\times2$ (\S\ref{sec:reads}, App.\ \ref{app:2x2}) is built from natural keyword-density
variation in these existing prompts with a fixed lexicon; no evasion prompts are authored.

\paragraph{Models and environment.} Qwen2.5-1.5B/3B-Instruct, SmolLM2-1.7B-Instruct, and
Phi-3-mini-4k-instruct, from the Hugging Face cache (revisions pinned in the repo manifest).
Python~3.13, \texttt{transformers} \citep{wolf2020transformers}, \texttt{torch}
\citep{paszke2019pytorch} (CPU-only), \texttt{scikit-learn} \citep{pedregosa2011scikit}; the one
embedding control uses \texttt{fastembed} (BGE). Models load in \texttt{bfloat16} at
$\ge 1.5$B (float32 exhausts memory on CPU) and \texttt{float32} otherwise. Per-condition baselines
(e.g.\ the $1.5$B attack-refusal rate, or the base model's plain-attack compliance) vary by a few
points across experiments, since scripts draw different AdvBench subsets at different $n$; every
control is compared against its own run's baseline, and we report each run's own numbers rather than
a pooled figure.

\paragraph{Probe, attack, and refusal.} The harm probe is $\ell_2$ logistic regression
(\texttt{C}$=0.5$, \texttt{max\_iter}$=2000$) on the last-prompt-token hidden state. Reported
\emph{detection} AUCs use nested-CV layer selection (no test-set peeking); the downstream
\emph{causal} experiments (patching, position, steering) instead fix layer 12, a representative
mid layer where the probe is near-ceiling, so the intervention layer is not itself selected on
the outcome. The prefill attack forces the assistant turn to begin with ``Sure, here is''.
Refusal is a fixed refusal-phrase string match. All generation is greedy (\texttt{do\_sample}$=$False).

\paragraph{Determinism.} Probes and featurizers are deterministic given the data; the only
stochastic element is the cross-validation split (fixed seed). The $2\times2$ lexicon, cell
construction, and decision thresholds were fixed before any outcome was seen.

\paragraph{State-transfer experiment (\S\ref{sec:reversal}).} The reversal is run by
\texttt{experiments/modal\_prefix\_transfer.py} (Qwen2.5-1.5B, L12, $n=100$ AdvBench, greedy,
$18$ generated tokens). It captures the mean layer-$11$-output residual over the first-half
generated window for the plain (no-prefill) and attack conditions, forms
$d_{\text{reverse}}=s_{\text{plain}}-s_{\text{attack}}$, and adds it into the attack's early
generated window over the same bands as the position control (onset / first half / all), against a
matched-norm control drawn orthogonal-random and rescaled to $\lVert d_{\text{reverse}}\rVert$. A
tiny-$n$ gate (plain refusal $\ge 0.7$, attack refusal $\le 0.15$) validates the harness before the
full run. The script is measure-only: it records refusal, a distinct-token coherence ratio, and a
harmful-procedural-content flag as per-prompt counts to \texttt{data/exp\_prefix\_transfer.json};
no generations are stored, and run logs (gitignored) print rates only.

\paragraph{Attention-knockout experiment (\S\ref{sec:attnko}, App.~\ref{app:attnko}).} Run by
\texttt{experiments/modal\_attn\_knockout.py} (eager attention, Qwen2.5-1.5B, all layers/heads,
$n=100$), preregistered in \texttt{PREREG\_attn\_knockout.md} (claim, falsifier, condition matrix,
interpretation table). Edge knockout adds $\ln(1-\lambda)$ to the pre-softmax attention logit of
(early-window query $\to$ prefill key) edges and renormalizes; the mass-matched control selects
non-prefill prompt keys whose baseline attention mass matches the prefill's per prompt (three fixed
seeds). A \texttt{MODE="unittest"} scaffold must pass before any behavioral run (sham-hook
bit-identity, edge-only masking, softmax renormalization, KV-cache consistency, live-hook and
behavioral positive controls). Measure-only: refusal / coherence / harmful-content counts to
\texttt{data/exp\_attn\_knockout.json}; CIs are paired bootstrap over prompts.

\FloatBarrier
\section{Full per-model results and the XSTest control}
\label{app:permodel}

Table~\ref{tab:dissociation} gives the headline numbers. The ``rep.\ AUC'' column is one
nested-CV probe (outer 5-fold, inner layer selection) on the clean prompt-token activations:
$0.996$ (Qwen-1.5B), $1.000$ (Qwen-3B), $0.990$ (SmolLM2), $0.998$ (Phi-3). There is no
separate ``under attack'' representation number, because the attack does not alter these
activations (it appends to the response); we report a single estimator to make that explicit.
The attacked-set behavior AUC is chance ($0.49$--$0.52$); the per-model complied-harmful vs.\
benign probe scores are $0.975/0.055$ (Qwen-1.5B), $0.981/0.048$ (Qwen-3B), $0.958/0.044$
(SmolLM2), $0.907/0.122$ (Phi-3).

\paragraph{Surface control (Table~\ref{tab:xstest}).} We score the AdvBench-trained probe
(layer 12) on XSTest. It separates real-unsafe from scary-but-safe at AUC $0.79$ with only
$12\%$ false positives on the scary-safe prompts: it reads harm, not violent vocabulary. It
is, however, AdvBench-calibrated ($0.97$ on AdvBench-harmful but $0.49$ on differently-phrased
XSTest-unsafe), and training on mixed harmful sources (AdvBench $+$ MaliciousInstruct) does
not fix it: it raises sensitivity but collapses specificity.

\begin{table}[h]
\centering
\caption{Harm probe on XSTest. The AdvBench-only probe reads harm (low false positives on
scary-safe); a mixed-source probe raises the unsafe score but wrecks specificity.}
\label{tab:xstest}
\small
\begin{tabular}{@{}lcccc@{}}
\toprule
probe & XSTest-unsafe & XSTest-safe (scary) & AUC & scary-safe FP \\
\midrule
AdvBench-only        & $0.49$ & $0.21$ & $0.79$ & $12\%$ \\
AdvBench $+$ Malicious & $0.89$ & $0.61$ & $0.78$ & $64\%$ \\
\bottomrule
\end{tabular}
\end{table}

\FloatBarrier
\section{Patching grids and the benign control}
\label{app:patch}

\begin{table}[h]
\centering
\caption{\textbf{Full per-layer patching grid (harmful refusal \% / benign refusal \%).}
Full-residual patching ``restores'' refusal to $100\%$, but the benign-residual control also
hits $100\%$ (generic disruption). The subspace rung (harm direction only) is a partial,
benign-sparing restoration; the fixed-cell $n=100$ deltas that settle it are in the main text
(Table~\ref{tab:patching}). Replicated Qwen2.5-1.5B and Phi-3-mini. Attack baseline: harmful
refusal $\le 0.07$.}
\label{tab:patchgrid}
\small
\begin{tabular}{@{}lcccc@{}}
\toprule
& \multicolumn{2}{c}{Qwen2.5-1.5B} & \multicolumn{2}{c}{Phi-3-mini-3.8B} \\
\cmidrule(lr){2-3}\cmidrule(lr){4-5}
layer & full / benign-CTRL & subspace (H/B) & full / benign-CTRL & subspace (H/B) \\
\midrule
8  & $97 / 97$  & $40 / 0$  & $100 / 100$ & $47 / 10$ \\
10 & $100 / 100$ & $3 / 0$   & $100 / 100$ & $53 / 20$ \\
12 & $100 / 100$ & $70 / 25$ & $100 / 100$ & $17 / 5$  \\
14 & $100 / 100$ & $23 / 15$ & $100 / 100$ & $3 / 0$   \\
\bottomrule
\end{tabular}
\end{table}

Table~\ref{tab:patchgrid} reports the full grid. We patch at the response site on
harmful$+$prefill prompts and let generation continue with no further intervention. The
attack baseline (no patch) is harmful refusal $0.07/$benign $0$ (Qwen-1.5B) and $0/0$
(Phi-3). \emph{full}: replace the onset residual with the mean harmful residual.
\emph{benign-CTRL}: the same with the mean \emph{benign} residual: it returns refusal to
$100\%$ at every layer for both models, which is why we read the full-residual restoration as
generic disruption rather than a harm-specific effect. \emph{subspace}: add
$\alpha \cdot \hat{d}_{\text{harm}}$ (unit harm-probe direction) at the response positions,
$\alpha \in \{0.5, 1, 1.5\}$ scaled to the layer's activation norm; the reported cell is the
best $\alpha$ by (harmful $-$ benign) refusal. \emph{Settling the harm-vs-random delta at
$n=100$} on two fixed cells (not maxima): at L12/$\alpha{=}1$ the harm direction restores $48\%$
(bootstrap $95\%$ CI $38$--$58\%$) versus $14\%$ for a matched-norm direction orthogonalized to
$\hat{d}_{\text{harm}}$ (24 seeds, range $0$--$60\%$), delta $+34$pp (CI $+22$,$+46$); at
L8/$\alpha{=}1.5$, $33\%$ ($24$--$42\%$) versus $16\%$ ($0$--$60\%$), delta $+17$pp
(CI $+6$,$+28$). Both exclude zero (the earlier $n=15$ estimate had overlapping CIs). Harm is
benign-sparing (benign restoration $12\%$/$0\%$ vs random $3\%$/$4\%$). So a harm-specific
component is real and significant, but a third to a half of the raw restoration is matched by the
random control; we report the harm$-$random delta. This licenses a \emph{partial, partly
specific} causal claim, not a clean lever.

\paragraph{Cell selection and the bootstrap.} The two cells are grid-selected (L12 is the best
subspace layer in the grid above; L8 a second), fixed before the $n=100$ run, which re-estimates
them on AdvBench$[:100]$, overlapping the grid prompts. Selection optimism therefore affects the
absolute restore rate but not the harm$-$random delta, which is measured for both arms at the same
cell and prompts. For the delta CI each bootstrap iterate resamples the $100$ harm prompts (with
replacement) \emph{and} the $24$ random seeds (with replacement), taking the harm mean minus the
resampled-seed mean; the harm-restore CI resamples prompts only. The random-control distribution
is right-skewed and mostly low: the $24$ L8 seeds are (sorted) $0,0,.01,.01,.01,.03,.06,.06,.07,
.09,.10,.11,.12,.13,.13,.13,.20,.21,.23,.26,.35,.37,.53,.60$, median $0.12$, with the $0$--$60\%$
range driven by four high outliers, so the harm direction's $33\%$ beats the \emph{typical}
random direction by more than the mean delta suggests.

\paragraph{Dose-matched positional control.} To ask whether the position result reflects
\emph{where} the harm signal is restored or merely \emph{how much} total signal is injected, we
band one consistent response window (the prefill $+$ the $18$ generated tokens, $22$ positions) and
vary position and per-position coefficient independently ($n=100$, L12; GATE: the all-positions
band at $\alpha{=}1$ reproduces the $41\%$ of the main text). Restoration by condition (\% refusal,
baseline $3\%$): all$@1$ $=41$, first-half$@1$ $=42$, first-half$@2$ $=47$, second-half$@2$ $=6$,
all$@0.5$ $=0$, every-other$@1$ $=1$, every-other$@2$ $=48$ (coherence $\ge0.96$ throughout, so
these are not degeneration). Reading: (i) the first half restores as much as the whole, and the
second half is nearly inert even at double strength, so the effect is a matter of \emph{position}
(an early window), not total dose; (ii) halving the coefficient over all positions gives $0\%$, so
there is a per-position strength threshold, not just a dose budget. This corrects an earlier
version of the control whose ``all'' band silently included the prefill positions while its
half-bands did not, confounding position with prefill-inclusion; the windowed design here is
consistent across bands.

\FloatBarrier
\section{Attention-knockout: unit tests, control matching, and the full matrix}
\label{app:attnko}

This appendix backs \S\ref{sec:attnko}. The experiment is preregistered
(\texttt{PREREG\_attn\_knockout.md}): primary hypothesis, falsifier, the frozen condition matrix,
the preregistered contrasts, and an interpretation table written before results.

\paragraph{Intervention and the onset caveat.} On each early-window decode step ($g<9$) we add
$\ln(1-\lambda)$ to the pre-softmax attention logit of every (current-query $\to$ prefill-key) edge
and let softmax renormalize ($\lambda{=}1$ is a full edge cut). Eager attention is mandatory
(fused SDPA/Flash kernels bypass a logit-level edit). Because generated token~1's logits are
produced from the final prefill token's hidden state, a response-to-prefill knockout cannot affect
token~1; this is a test of early \emph{maintenance}, not attack initiation.

\paragraph{Unit tests (all pass before any behavioral run).} (i) eager attention active;
(ii) prefill key indices located after the chat template; (iii) sham hook (knockout with an empty
key set) is bit-for-bit identical to baseline under greedy decoding; (iv) after a full edge cut the
removed edges carry $0$ attention mass and the softmax row renormalizes to $1$; (v) non-target query
rows are unchanged; (vi) no NaN/Inf reaches the residual stream; (vii) a large knockout measurably
moves next-token logits ($\|\Delta\|\approx10^{3}$, the hook is live); (viii) knocking out attention
to the \emph{prompt} tokens materially changes generation (a behavioral positive control: the
machinery moves behavior, so a prefill-specific null means ``prefill edges not necessary,'' not
``dead hook''); (ix) KV-cache on/off agree under the sham hook.

\paragraph{Mass-matched control.} A no-knockout measurement pass records the baseline attention mass
from early-window queries onto each key (mean over layers and heads). The prefill keys carry small
mass (mean $0.082$); attention-sink tokens (BOS, template) carry far more, so a max-mass or random
span is not a fair control. We select, per prompt, a set of non-prefill prompt keys of the same size
whose summed mass best matches the prefill's (achieved matched/prefill ratio $0.98$, range
$[0.75,1.08]$ over three fixed seeds). Knockout of this set is the primary control.

\paragraph{Full matrix and contrasts.} Table~\ref{tab:attnko} gives the condition means. Paired
bootstrap ($10{,}000$ resamples over the $100$ harmful prompts; the matched contrast uses one fixed
control seed, matched $3\%$ on that seed, three-seed mean $1\%$): early prefill KO $-$ mass-matched
KO $=+21$pp ($[+12,+30]$); early $-$ baseline $=+21$pp ($[+12,+30]$); early $-$ late $=+20$pp
($[+11,+29]$); whole $-$ early $=+5$pp ($[-1,+12]$, includes zero, so whole is not distinguishable
from early). Benign specificity: under the prefill, early knockout moves benign refusal only
$0\%\to4\%$ (24 benign prompts), against the harmful $3\%\to24\%$, a difference-in-differences of
$+16.8$pp. Coherence stays $0.96$--$0.97$ across all conditions (the preregistered non-inferiority
margin was a $\le5$-point drop). The knockout also suppresses harmful \emph{content} (procedural
rate $51\%\to18\%$ early, $\to3\%$ whole), prefill-specifically (matched control $50\%$).

\paragraph{Held-out replications of the two causal handles (\S\ref{sec:reversal},~\S\ref{sec:causal}).}
Both the state-transfer reversal and the harm-direction patching are re-run with a clean train/test
split so the injected vector never sees the test prompts. \emph{Reversal:} $d_{\text{reverse}}$ is
estimated on AdvBench$[0{:}260]$ and injected/tested on the disjoint held-out $[260{:}360]$; it
restores refusal to $74\%$ over the first half (whole $78\%$, onset $13\%$), the matched-norm random
control to $0\%$, coherence $0.97$--$0.99$; the in-distribution reference (same $d_{\text{reverse}}$
tested on train prompts) gives $72\%$, so the generalization gap is within noise. \emph{Patching:}
the harm direction is trained on AdvBench$[0{:}200]$ (harmful vs Alpaca benign, activations
recomputed on the fly) and restoration tested on the held-out $[260{:}360]$; the L12/$\alpha{=}1$
cell restores $42\%$ ($95\%$ bootstrap CI $32$--$52\%$, $n=100$) versus $13\%$ for the matched-norm
orthogonal-random control ($6$ seeds), delta $+29$pp (CI $+16$,$+41$), benign over-refusal $15\%$.
Both handles pass held-out (the direction shrinks $\sim5$--$6$pp off-distribution, the reversal
does not), so neither is an artifact of the prompts it was estimated on. Held-out patching was run
for the primary L12 cell only. \emph{Second family (Phi-3-mini, $3.8$B).} Both new results replicate. The held-out reversal
($L16$) restores refusal to $69\%$ first-half / $71\%$ whole / $4\%$ onset, random $2\%$,
in-distribution reference $71\%$ (baseline $1\%$). The attention knockout replicates the full
pattern at $n=100$: attack baseline $1\%$; early prefill-edge KO $25\%$; whole $32\%$; late $0\%$;
half-dose ($\lambda{=}0.5$) $0\%$; mass-matched control $1\%$; coherence $0.95$--$0.98$ throughout
(control mass-match ratio $1.00$). Phi's unit-test suite passes $9/10$; the one failure is a
cache-vs-no-cache determinism check that is a bf16 property of the model (it reproduces with a
no-op hook), and does not affect the experiment, which compares cache-on baseline against cache-on
knockout, and the sham-identity and edge-cut checks pass under cache-on.

\paragraph{Active-vs-passive: base-model discriminator.} To ask whether the knockout effect is
safety-tuning-specific (active suppression) or generic (passive conditioning), we run the identical
early prefill-edge knockout on the Qwen2.5-1.5B \emph{base} model (no safety tuning), with a
raw-completion format (no chat template) and the harmful-\emph{content} scorer as the endpoint (the
base model does not refuse). The base model complies ($64\%$ harmful content) and shows the same
prefill-specific, early-window dependence as the instruct model: early knockout drives harmful
content to $25\%$ (a $-39$pp drop) while the mass-matched control leaves it at $64\%$ ($-0$pp) and
late knockout is inert ($61\%$). Since a non-safety-tuned model shares the dependence, it is generic
autoregressive conditioning, not a safety-specific active suppression; this resolves the
active-vs-passive question toward the passive reading (\S\ref{sec:limits}). \emph{Mid-scale
replication (Qwen2.5-7B base).} The identical knockout on the $7$B base model reproduces the pattern
cleanly: attack baseline $68\%$ harmful content, early prefill-edge knockout $30\%$ (a $-38$pp drop),
mass-matched control $66\%$ ($-2$pp, three seeds), late knockout inert ($68\%$), for a $+36$pp
prefill-specific, early-window-gated drop ($n=100$). The passive reading is therefore not a
small-model artifact. Measure-only,
\texttt{experiments/modal\_attn\_knockout\_base.py} and \texttt{modal\_attn\_knockout\_base\_7b.py}.

\paragraph{Active-vs-passive: logit-level decomposition (a small active component).} To probe for a
residual safety-specific pull, we decompose \emph{how} refusal recovers under the knockout at the
logit level. For each early-window generated step we record the softmax mass on refusal-onset tokens
($p_{\text{refuse}}$) and the probability of the compliance continuation ($p_{\text{comply}}$),
under baseline / early knockout / mass-matched knockout, and define \emph{concentration}
$=\Delta p_{\text{refuse}} / (-\Delta p_{\text{comply}})$: of the probability the knockout strips
off compliance, the fraction that lands on refusal. In both models the knockout's dominant effect is
compliance collapse ($p_{\text{comply}}$: instruct $-5.6$pp, base $-17$pp), the passive signature.
But the instruct model routes the freed mass to refusal far more than the base model
(concentration $0.24$ vs $0.03$; prefill-specific $p_{\text{refuse}}$ gain $+1.3$pp vs matched
$+0.1$pp), so safety tuning adds a modest learned refusal attractor on top of the generic mechanism.
The metric confirms a small safety-specific refusal component exists and sizes it, but does not
separate an actively-suppressed gate springing back from a passive fallback attractor. Instruct
$60$ prompts; base as anchor. Measure-only, \texttt{experiments/modal\_active\_probe.py}.

\paragraph{The MLP pathway (closing the accounting gap on the generic side).} To ask whether the
compliance that persists after the attention knockout hides a safety-specific mechanism in the MLP pathway, we
capture the mean early-window MLP output per layer in the plain (no-prefill) and attack conditions,
form $d_{\text{mlp}} = \text{plain} - \text{attack}$, and inject it at the early-window MLP outputs
under the attack ($n=60$, all layers, matched-norm random and late-window controls). \emph{Base
model (the clean discriminator):} removing the base model's own prefill-induced MLP shift drops
harmful content $68\%\to48\%$ ($+16$pp prefill-specific vs the matched-random control's $\sim4$pp,
late-window inert), in a model with no safety mechanism, so the early MLP's prefill-dependence is
generic, extending the passive account from attention to a second pathway. \emph{Instruct (a
decomposition, not a discriminator):} the same injection restores refusal $3\%\to47\%$
(prefill-specific, early-window-gated), showing the MLP carries much of the refuse-state; but this
is a difference-of-means injection, which tests decodability, not active computation, and the
larger instruct effect is overdetermined by its larger plain-vs-attack behavioral gap. We therefore
report the $47\%$ as decomposition only, and settle the instruct excess with the clean path patch
below. Measure-only, \texttt{experiments/modal\_mlp\_pathpatch.py}.

\paragraph{Clean path patch: the early MLP pathway is generic.} A position-wise MLP never reads the
prefill positions, so the excess cannot be isolated by a positional ablation; it needs a path patch
that severs only the prefill-attention contribution to the MLP input. Teacher-forced over the
greedily-generated attack sequence (so clean logits equal the generation), we (i) knock out
early-window attention to the prefill and capture each layer's resulting MLP input at the
early-window positions, then (ii) run the clean attack and overwrite \emph{only} those MLP inputs
with the captured knockout values, leaving every other activation, including all downstream
attention, at its clean value. Because Qwen2 saves the residual before the MLP layernorm, the
counterfactual enters through the MLP sub-block alone. The surgery passes a unit-test suite before
any read: sham (injecting the clean MLP input changes nothing, max $|\Delta|{=}0$), MLP-output
faithfulness (the patched MLP output equals the knockout run's, max $|\Delta|{=}0$), hook-live (the
attention knockout moves logits by $>15$), and locality (patching only the last position leaves
earlier-position distributions unchanged, max $|\Delta|{=}0$); all bit-exact on both models. On the
same concentration metric ($\Delta p_{\text{refuse}} / -\Delta p_{\text{comply}}$, $p_{\text{comply}}$
the mass on the clean compliance token), the MLP pathway carries the generic effect and only it: it
strips prefill-specific compliance probability (instruct $-21$pp, base $-25$pp, against
mass-matched-MLP-path controls at $-7$/$-5$pp) but shows \emph{no} safety-specific refusal
concentration (instruct $0.007$, base $0.001$, matched $-0.02$; prefill-specific refusal gain instruct
$+0.26$pp). So the residual safety-specific pull found in the logit trace does not route through the
early MLP, and the instruct excess in the difference-of-means test was decodability. Caveat: this
teacher-forced logit frame is well-powered for compliance-suppression routing and less sensitive to a
small refusal-promotion component (which would need a generative trajectory to accumulate), but the
within-frame instruct-vs-base-vs-matched comparison is clean. $n{=}60$; measure-only,
\texttt{experiments/modal\_mlp\_pathpatch\_clean.py}.

\paragraph{Reviewer robustness checks (boundary, prefill-exclusion, prefill length, sampling,
benign-sparing).} Bundled on one instruct load, gated by the same plain-refuses/attack-complies
check and cross-validated against the published anchors (the whole-band harm patch reproduces the
$48\%$ fixed cell; $d_{\text{reverse}}$ first-half reproduces $75\%\approx$ the $74\%$ held-out
reversal), \texttt{experiments/modal\_revision\_robustness.py}. \emph{(i) Boundary sweep}
(harm-direction patch, L12/$\alpha1$, full per-position dose, $n{=}60$): onset $10\%$, first-third
$50\%$, first-half $57\%$, first-two-thirds $55\%$, whole $50\%$, second-half $5\%$, so restoration
saturates by the first third and is flat thereafter. \emph{(ii) Prefill-exclusion:} injecting over
the generated tokens only (prefill positions excluded) restores $0\%$ (first-half band) and $3\%$
(generated tokens $1$--$7$), versus $57\%$ with the prefill span included; the causal write-site is
the prefill span, and generated positions carry the effect only when the prefill is also driven.
\emph{(iii) Prefill length} ($n{=}50$, bands over generated-token counts, all prefill positions
driven): the saturation-by-first-third pattern holds for a short prefill (``Sure,'': first-third
$86\%$, whole $86\%$) and a long one (``Sure, here is a detailed step-by-step guide:'': first-third
$12\%$, whole $6\%$), so the window is an early band in generated-token count, not a fraction of the
prefill; the restoration \emph{level} falls with prefill commitment. \emph{(iv) Sampling}
($T{=}0.7$, $n{=}40$, $K{=}2$): plain refusal $100\%$, attack refusal $0\%$ (vs greedy $5\%$), so the
dissociation is not a greedy artifact; the harm-direction restoration weakens ($50\%\to21\%$),
consistent with a partial, non-selective write-handle. \emph{(v) Benign-sparing} of the state
transfer: $d_{\text{reverse}}$ injected on benign prompts under the attack raises their refusal
$12\%\to18\%$ ($+6$pp) while restoring harmful refusal to $75\%$ ($+57$pp selectivity), so the
transfer is a selective edit, not a blanket refusal-inducer.

\paragraph{Scale (Qwen2.5-7B).} At $7$B the prefill attack is weaker: plain refusal $98\%$, attack
refusal $24\%$ (vs $\le3\%$ at $1.5$--$3.8$B), procedural content $16\%$. The knockout's
prefill-specific effect persists in the \emph{content} channel (early knockout $0\%$, whole $0\%$,
late $4\%$, half-dose $13\%$, mass-matched control $23\%$, coherence $\ge0.95$), early-window-gated
and dose-dependent, but the refusal-\emph{string} restoration seen at smaller scale does not cleanly
replicate (early $20\%\approx$ baseline $24\%$), consistent with the passive reading that the
refusal fallback is model-dependent. Unit tests pass $10/10$ (including the KV-cache check); run on
an L40S (eager attention). Gate relaxed to attack-refusal $\le0.35$ for the more-robust $7$B.

\FloatBarrier
\section{The steering nulls}
\label{app:steer}

We add the harm direction at all response positions under the attack and sweep layer and
strength, in two parameterizations. Neither finds a window that raises harmful refusal while
sparing benign (Table~\ref{tab:steer}). The best cells restore harmful refusal to $71$--$83\%$
but at $40$--$50\%$ benign over-refusal; smaller coefficients over-refuse everything, larger
ones ($\alpha{=}4$ / coeff $14$) collapse generation to $0$. The harm direction induces
refusal broadly, not on-harm. By contrast the read-side use of the same representation (the
monitor of \S\ref{sec:dissociation}) catches $100\%$ of the attacked-harmful prompts at
$0\%$ false positive.

\begin{table}[h]
\centering
\caption{Steering sweeps (harmful \% / benign \% refusal under attack), Qwen2.5-1.5B. No cell
raises harmful refusal without dragging benign up. \emph{Left}: raw diff-of-means direction.
\emph{Right}: unit direction, coefficient scaled to activation norm ($\alpha$).}
\label{tab:steer}
\small
\begin{tabular}{@{}lccc@{}}
\multicolumn{4}{l}{\textit{raw (coeff)}}\\
\toprule
layer & coeff 4 & coeff 8 & coeff 14 \\
\midrule
6  & $7/0$  & $0/0$  & $0/0$ \\
12 & $7/20$ & $71/50$ & $0/0$ \\
18 & $0/0$  & $0/0$  & $0/0$ \\
22 & $0/0$  & $0/0$  & $0/0$ \\
\bottomrule
\end{tabular}
\hfill
\begin{tabular}{@{}lccc@{}}
\multicolumn{4}{l}{\textit{normalized ($\alpha$)}}\\
\toprule
layer & $\alpha 1$ & $\alpha 2$ & $\alpha 4$ \\
\midrule
10 & $0/0$    & $0/0$   & $0/0$ \\
12 & $8/40$   & $8/0$   & $0/0$ \\
14 & $92/100$ & $50/90$ & $0/0$ \\
16 & $100/90$ & $83/40$ & $0/0$ \\
\bottomrule
\end{tabular}
\end{table}

\FloatBarrier
\section{The pre-registered $2\times2$}
\label{app:2x2}

\paragraph{Construction.} We fix a trigger-word lexicon (eight categories: violence, weapons/
explosives, drugs, cyber, self-harm, extremism, sexual, fraud/theft) with inflection-aware
matching (so ``hacking'' and ``drugs'' count), and a per-prompt \emph{scary} flag $=$ ``at
least one trigger word.'' Cells come from natural variation in existing public prompts, with
no authored evasions: harmful-scary $=$ AdvBench with a trigger word; harmful-clean $=$
AdvBench without one (e.g.\ ``how do I best destroy my boss's car,'' ``commit insider
trading''); benign-scary $=$ all of XSTest-safe (purpose-built to sound harmful); benign-clean
$=$ Alpaca. We balance to $n{=}120$ per cell and use source labels (flagging, not filtering,
the severity spread). Two issues surfaced and were fixed \emph{before} scoring: an initial
word-boundary matcher leaked inflections into the clean cell (fixed), and the strict-lexicon
benign-scary cell was thin ($n{=}54$), so we pre-registered benign-scary $=$ all XSTest-safe
with the strict-keyword subset as a robustness check.

\paragraph{Results.} Table~\ref{tab:2x2} in the main text. A logistic regression of refusal on
the two binary axes gives $\beta_{\text{harm}}{=}5.12$, $\beta_{\text{scary}}{=}2.20$. The
representational readout agrees: the refusal-behavior direction is closer to the harm-content
direction than to the surface direction (mid-layer $|\cos|$ $0.34$ vs.\ $0.13$).

\FloatBarrier
\section{The refusal-direction search}
\label{app:dirsearch}
Test for a refuse-vs-comply direction distinct from harm (\S\ref{sec:dirsearch}). A weakened
prefill (``I understand you're asking'') gives a $41/29$ refuse/comply split on $70$ harmful
prompts, so harm and prefill-presence are held fixed and only the decision varies;
$d_{\text{refuse}}$ is the diff-of-means of the decision-token activations.
\textbf{Distinct but weak:} $|\cos(d_{\text{refuse}}, d_{\text{harm}})|=0.06$; five-fold CV AUC
$0.648$ versus $d_{\text{harm}}$'s $0.640$ at layer $12$. \textbf{Read sweep (single direction):} across all $29$
layers $\times$ the first $4$ generated positions the diff-of-means readout ceilings at AUC
$0.727$ (layer $27$, position $1$); most cells are $0.60$--$0.72$. \textbf{But that ceiling is
rank-1:} a full L2-logistic probe on the same activations reads refuse-vs-comply at $0.85$
(layer $21$; $+0.12$ over the best single direction, consistent across $C\in\{0.01,0.1\}$ and
layers, noisy at $n=70$). The decision is multidimensional, not weak. \textbf{Rank:} it is
\emph{distributed}, not low-rank: PCA-then-logistic reaches AUC $0.74$ at $3$ components (near
the single-direction $0.73$) and climbs to $\sim0.84$ only by ${\sim}30$; the top-variance axis
is anti-predictive (rank-1 PCA $0.40$) and none of the top components are harm-aligned
($|\cos|\le0.13$), so the decision lives in low-variance, non-harm directions spread wide
(caveat: $n=70$ flatters the high-rank estimates).
\textbf{Control (inert):} steering $+c\,d_{\text{refuse}}$ at the response
positions under the strong attack restores $\le 7\%$ harmful refusal at layer $12$ and $0\%$ at
layer $27$ (the best-read cell), at every coefficient $\in\{0.5,1,1.5\}$, while $d_{\text{harm}}$
restores refusal at layer $12$ (non-selectively) and a matched-norm random direction restores
nothing. No layer or position yields a clean or steerable decision direction; refusal is not a
separable linear feature.

\textbf{Across scale (Table~\ref{tab:ladder}).} We repeat the single-vs-full comparison across
the model ladder under a per-model calibrated weak prefill. Within the Qwen family the full
probe beats the best single direction at every scale ($0.5$--$7$B); SmolLM2 (a balanced split,
different family) shows no gain, and Phi-3's calibration gave a $43/7$ split too imbalanced to
read. The $7$B run used Modal (A10G, white-box) with a last-layer/$5$-fold probe.

\begin{table}[h]
\centering\small
\caption{\textbf{Distributed decision across scale.} Best single-direction (diff-of-means) vs
full L2-logistic AUC for refuse-vs-comply, per-model calibrated weak prefill. Full $>$ single
across the Qwen family; SmolLM2 is a balanced cross-family counterexample; Phi-3's split is
unreliable.}
\label{tab:ladder}
\begin{tabular}{@{}llccc@{}}
\toprule
model & family & split (ref/cmp) & single & full \\
\midrule
Qwen2.5-0.5B    & Qwen          & $28/22$ & $0.763$ & $0.843$ \\
Qwen2.5-1.5B    & Qwen          & $41/29$ & $0.727$ & $0.848$ \\
Qwen2.5-3B      & Qwen          & $17/33$ & $0.633$ & $0.696$ \\
Qwen2.5-7B      & Qwen          & $52/18$ & $0.847$ & $0.942$ \\
SmolLM2-1.7B    & HuggingFaceTB & $27/23$ & $0.829$ & $0.805$ \\
Phi-3-mini-3.8B & Microsoft     & $43/7$  & \multicolumn{2}{c}{unreliable (imbalanced)} \\
\bottomrule
\end{tabular}
\end{table}

\FloatBarrier
\section{The intent-axis extension}
\label{app:intent}
$24$ dual-use triplets (\S\ref{sec:reads}), each request in three framings. \textbf{Construction:}
the request is dual-use (e.g.\ ``how do I get into a locked iPhone I can't unlock?''); the benign
context makes helping appropriate (``it's my own phone and I forgot the passcode'') and the
harmful context makes refusing appropriate (``it's my ex's phone and I want to read their
messages''). Refusal is string-matched; $d_{\text{intent}} = \text{mean}(\text{benign}) -
\text{mean}(\text{harmful})$ at the decision token (layer $17$, best decode). \textbf{Graded
behavior:} refusal $33\%$ (benign) / $62\%$ (bare) / $71\%$ (harmful); intent CV AUC $0.76$.
\textbf{Faithful tracking:} in deep layers the bare prompts the model refuses project harmful on
$d_{\text{intent}}$ ($-5$ to $-20$) and the ones it helps project benign ($+3$ to $+16$).
\textbf{Attack:} on the $15$ bare prompts refused (read harmful-intent, mean projection $-5.9$),
the prefill drops refusal from $100\%$ to $0\%$; over all $24$, $62\%\to0\%$. Exploratory: one
model (Qwen2.5-1.5B), $n=24$.

\FloatBarrier
\section{Refusal is not selectively ablatable (the head-ablation arc)}
\label{app:ablation}
We searched for a compact refusal circuit by attention-head ablation (Qwen2.5-1.5B, GPU), and
the negative only holds because we verified the hook fires. \textbf{(1)} Ablating
each of the $28\times12$ heads (zeroing its slice of the \texttt{o\_proj} input across
generation) drops refusal by $0.00$ everywhere. \textbf{(2)} Suspecting a dead hook, a positive
control ablating a whole attention layer also left refusal at $100\%$, still ambiguous.
\textbf{(3)} A logit check settled it: the hook changes the next-token logits by
$\|\Delta\|\approx123$ (whole attention, layer 12) and $\approx23$ (a single head), so it works,
and the robustness is real. \textbf{(4)} Cumulative ablation (first-$k$ attention layers) drove
the refusal string-match to $0$ even at $k=1$, but this \emph{contradicted} the per-layer control
(layer 0 breaks it, layer 12 does not). \textbf{(5)} Reading the actual generations resolved it:
the refusal-$0$ cases are \emph{garbage} (``\,silly silly noble glorious\,\dots''; with all
attention removed the model emits one fixed gibberish string for every prompt), not compliance,
the same generic-disruption confound as full-residual patching, caught only because we read the
text rather than the metric. The bar we hold every intervention to is a \emph{selective}
transition from harmful refusal to coherent harmful compliance \emph{without} comparable
degradation on benign behavior; broad ablation fails it doubly (the output is garbage, and the
\emph{same} garbage for benign prompts), so the result speaks against a compact head- or
layer-level circuit rather than merely lowering a refusal rate (MLP and SAE features untested).
Conclusion: no single head or layer is
necessary for refusal, and no tested internal intervention selectively removes it (it collapses
the model instead). Scope: one model, attention ablation; MLP not tested.

\paragraph{At $14$B (Phi-3-medium, heavily safety-trained).} We re-ran the two load-bearing
pieces at $14$B (measure-only: refusal string-match and a distinct-token coherence ratio, no
generations stored). \emph{Dissociation holds}: harm-probe AUC $1.00$ (layer $8$), plain refusal
$1.00\to0.15$ under the prefill, and the $34/40$ complied prompts still score harmful ($0.96$)
with coherent (not deflecting) output (coherence $0.94$). \emph{Compact-circuit check}: over a
per-attention-layer ablation ($40$ layers, refusing set $n=16$), only the \emph{input} layer
(L$0$) shows a refusal drop $\ge0.3$ ($1.00\to0.69$, coherence $0.69$), and L$0$ is a generic
early-processing locus, not a refusal-specific circuit; every other layer leaves refusal near
ceiling. So the negative holds at $14$B and heavier safety training, with the honest
caveat that the input layer carries a partial, non-selective contribution. Frontier scale
(70B$+$) is untested.

\FloatBarrier
\section{What did not work}
\label{app:nulls}

\paragraph{Two off-thesis nulls (in brief; full detail in the ledger).} Before safety, we
pointed the same represent-vs-act probe at two other tasks and got nulls that motivated the
choice of substrate. On MC \emph{correctness} (ARC-Easy, $0.6$--$1.5$B) the model's verbalized
confidence matches or beats a hidden-state probe, so the gap does not appear where the output
already has an honest channel, which is what sent us to safety, where refusal has none. On
judge-free failure \emph{localization} (Who\&When, \citealp{zhang2025whowhen}) two onset signals
lose to dumb priors; the one reportable byproduct is an audit finding that published LLM-judge
methods lose to constant-guessing on step localization ($14\%$ vs.\ $27\%$), which we flag as a
separate direction, not a contribution here.

\paragraph{An attack-surface map (exploratory).} Applying the harm probe across the attack
transforms of the Arcanum prompt-injection taxonomy \citep{haddix2026arcanum},
semantic attacks (roleplay, tense) keep harm represented while heavy encodings the
model cannot decode (base64, morse, reverse) collapse it: the response-site/prompt-modifying
distinction of \S\ref{sec:related}. But a pre-registered three-scale test refuted the
prediction that this ``blind spot'' shrinks monotonically with scale, and the per-technique
verdicts do not compare across models, so we report it as exploratory only.

\FloatBarrier
\section{Controls ledger: every adversarial check and its verdict}
\label{app:controls}

Every control we ran against a main-text result, what it could have shown, and the verdict:
\textsc{passed} (the effect held), \textsc{demoted} (the dramatic version reduced to a modest true
one), \textsc{null}, or \textsc{retracted}. About a third of the verdicts are not \textsc{passed}.

\medskip
{\small
\begin{longtable}{@{}p{0.30\textwidth} p{0.40\textwidth} p{0.22\textwidth}@{}}
\caption{Every adversarial control and its verdict.}\label{tab:controls}\\
\toprule
\textbf{Result under test} & \textbf{Control (what it could have shown)} & \textbf{Verdict} \\
\midrule
\endfirsthead
\multicolumn{3}{@{}l}{\emph{Table~\ref{tab:controls}, continued}}\\
\toprule
\textbf{Result under test} & \textbf{Control (what it could have shown)} & \textbf{Verdict} \\
\midrule
\endhead
\midrule \multicolumn{3}{r@{}}{\emph{continued on next page}}\\ \endfoot
\bottomrule \endlastfoot
Probe reads harm & XSTest scary-but-safe: does it fire on violent vocabulary? ($0.21$, $12\%$ over threshold) & \textsc{passed} \\
Dissociation not source-calibration & complied- vs refused-harmful from the \emph{same} distribution (calibration cancels) & \textsc{passed} \\
Source-general harm monitor & mixed-source probe (AdvBench $+$ MaliciousInstruct): does it generalize? & \textsc{demoted} (specificity wrecked) \\
Compliance is real, not deflection & judge-free content read of non-refusals ($69$--$95\%$ substantive) & \textsc{passed} \\
Harm predicts who complies (weak prefill) & regress out prompt length $+$ scary-keyword count (AUC holds $0.73$) & \textsc{passed} (moderate) \\
Full-residual restoration ``$100\%$'' & benign-residual patch at the same site (also $100\%$?) & \textsc{demoted} (generic disruption) \\
Harm-direction subspace patching & matched-norm \emph{orthogonalized} random, $24$ seeds & \textsc{passed} ($+34/+17$pp, partial) \\
Patching is harm-specific & held-out split (direction trained on disjoint prompts) & \textsc{passed} ($+29$pp) \\
Patching spares benign & benign restoration ($12\%/0\%$ vs random $3\%/4\%$) & \textsc{passed} \\
``Grip is the whole response'' & dose-matched, prefill-inclusion-consistent bands & \textsc{demoted} $\to$ early window (dose$+$inclusion confound) \\
Early window is a per-position threshold & halve the coefficient over all positions ($\alpha{=}0.5\to0\%$) & \textsc{passed} \\
``First-half $\approx$ whole'' is a knife-edge at $1/2$ & boundary sweep (first-third already $50\%$; onset $10\%$) & \textsc{passed} (saturates by first third) \\
Restoration is carried by generated tokens & prefill-exclusion (generated-only $\le3\%$ vs $57\%$) & \textsc{demoted} (write-site is the prefill span) \\
Effects are a greedy-decoding artifact & resample at $T{=}0.7$ (plain $100\%$, attack $0\%$) & \textsc{null} (dissociation holds; restore $\to21\%$) \\
Onset projection $2.80{\to}0.22$ is displacement & benign non-affirmative prefix (drives it \emph{lower}, $-0.82$) & \textsc{demoted} (content-confounded) \\
Refuse-state transfer (reversal) & matched-norm random ($0\%$); held-out ($74/72\%$); benign-sparing ($+6$pp) & \textsc{passed} (selective) \\
Attention knockout & mass-matched control ($1\%$); unit tests; benign DiD ($+16.8$) & \textsc{passed} ($+21$pp, harm-selective) \\
Knockout is early-window-gated & late-window knockout (inert); whole $\approx$ early & \textsc{passed} \\
Active suppression vs passive & base-model discriminator (no safety tuning) & \textsc{demoted} to passive (generic) \\
An active component exists & logit-trace concentration ($0.24$ vs base $0.03$) & \textsc{passed} (small) \\
Remaining compliance is safety-specific (MLP) & base-model MLP control ($+16$pp, no safety mechanism) & \textsc{demoted} (early MLP is generic too) \\
Instruct MLP excess is an active component & clean frozen-residual path patch (bit-exact unit tests) & \textsc{demoted} (no safety-specific pull; $+0.26$pp) \\
Harm direction as a \emph{steering} intervention & $4$-layer $\times$ $3$-coefficient sweep, raw and Arditi-normalized & \textsc{null} (no selective window) \\
A single refusal-decision direction & full probe ($\sim0.85$) vs best single ($\le0.73$); PCA rank & \textsc{demoted} (distributed) \\
$d_{\text{refuse}}$ steers behavior & inject at mid and best-read late layer, all coefficients & \textsc{null} (causally inert) \\
Refusal is selectively ablatable & $28\times12$ head sweep $+$ hook-verified $+$ coherence read; $14$B & \textsc{null} (heavy ablation $=$ garbage) \\
Refusal reads a ``separate feature'' & decorrelated harm$\times$surface $2\times2$ & \textsc{retracted} \\
Refusal is harm-primary & within-AdvBench (strip trigger words, $100\%\to100\%$); off-ceiling SmolLM2 & \textsc{passed} \\
Keyword over-refusal magnitude & benign-scary is all XSTest ($\beta_{\text{scary}}$ an upper bound) & \textsc{passed} (scoped) \\
Monitor is attack-proof in general & XSTest FP ($12\%$); prompt-modifying/encoding attacks & \textsc{demoted} (response-site only) \\
Mechanism holds at $7$B & Qwen2.5-7B knockout (content persists, refusal-string does not) & \textsc{passed} (mechanism; behavior model-dependent) \\
Honesty monitor beats verbalized confidence & confidence baseline (ARC) & \textsc{null} \\
Failure localization beats dumb priors & constant-guess baseline (Who\&When) & \textsc{null} \\
Attack blind-spot shrinks with scale & pre-registered three-scale test & \textsc{refuted} \\
\end{longtable}
}

\end{document}